\documentclass[journal]{IEEEtranTIE}
\usepackage{graphicx}
\usepackage{cite}
\usepackage{picinpar}
\usepackage{amsmath}
\usepackage{url}
\usepackage{flushend}
\usepackage[latin1]{inputenc}
\usepackage{colortbl}
\usepackage{soul}
\usepackage{multirow}
\usepackage{pifont}
\usepackage{color}
\usepackage{alltt}
\usepackage[hidelinks]{hyperref}
\usepackage{enumerate}
\usepackage{siunitx}
\usepackage{breakurl}
\usepackage{epstopdf}
\usepackage{pbox}
\usepackage{bm}
\usepackage{amssymb}
\usepackage{float}

\usepackage{textcomp}
\usepackage{stfloats}
\usepackage{changes}
\usepackage{subcaption}
\usepackage{url}
\usepackage{verbatim}
\usepackage{graphicx}
\usepackage{cite}
\usepackage{hyperref}

\usepackage{amsmath, amsfonts, amssymb, amsthm}
\usepackage{mathtools}
\usepackage{booktabs}
\usepackage{multirow}
\usepackage{multicol}
\usepackage[table]{xcolor}
\usepackage{makecell}

\usepackage{caption}
\makeatletter
\newcommand{\rmnum}[1]{\romannumeral #1}
\newcommand{\Rmnum}[1]{\uppercase\expandafter{\romannumeral #1}}
\makeatother

\begin{document}
\title{LOONG: Online Time-Optimal Autonomous Flight for MAVs in Cluttered Environments}


\author{
	\vskip 1em

        Xin Guan, 
        Fangguo Zhao,
        Qianyi Wang,
        Chengcheng Zhao,
	Jiming Chen, \emph{Fellow, IEEE},
	and Shuo Li, 

    

	\thanks{
	
        
        
        The authors are with the College of Control Science and Engineering, Zhejiang University, Hangzhou 310027, China (e-mail: shuo.li@zju.edu.cn) \textit{(Corresponding author: Shuo Li.)} . 
		
	}
}

\maketitle
	


\begin{abstract}

Autonomous flight of micro air vehicles (MAVs) in unknown, cluttered environments remains challenging for time-critical missions due to conservative maneuvering strategies. This article presents an integrated planning and control framework for high-speed, time-optimal autonomous flight of MAVs in cluttered environments. In each replanning cycle (100\,Hz), a time-optimal trajectory under polynomial presentation is generated as a reference, with the time-allocation process accelerated by imitation learning. Subsequently, a time-optimal model predictive contouring control (MPCC) incorporates safe flight corridor (SFC) constraints at variable horizon steps to enable aggressive yet safe maneuvering, while fully exploiting the MAV's dynamics. We validate the proposed framework extensively on a custom-built LiDAR-based MAV platform. Simulation results demonstrate superior aggressiveness compared to the state of the art, while real-world experiments achieve a peak speed of 18\,m/s in a cluttered environment and succeed in 10 consecutive trials from diverse start points. The video is available at the following link: https://youtu.be/vexXXhv99oQ.


\end{abstract}

\begin{IEEEkeywords}
Aerial robotics, obstacle avoidance, integrated planning and control, motion planning.
\end{IEEEkeywords}

{}

\definecolor{limegreen}{rgb}{0.2, 0.8, 0.2}
\definecolor{forestgreen}{rgb}{0.13, 0.55, 0.13}
\definecolor{greenhtml}{rgb}{0.0, 0.5, 0.0}

\section{Introduction}


\IEEEPARstart{A}{utonomous} navigation techniques~\cite{9143458},\cite{10649014} for micro air vehicles (MAVs) leverage onboard sensing and computation to perceive the unknown environment and plan collision-free trajectories. 
These capabilities have rendered autonomous MAVs
increasingly indispensable in diverse applications, including 
aerial photography
and industrial inspection~\cite{9900135}.
Their importance is further amplified in time-critical missions such as search and rescue~\cite{mishra2020drone}
and disaster relief~\cite{daud2022applications}, where timely and safe arrival 
is crucial. 
In parallel with these practical demands, autonomous MAVs have continuously broken speed records in recent years, driven by the emergence of autonomous drone 
racing~\cite{ moon2019challenges},\cite{hanover2024autonomous}.
This competitive platform, with its emphasis on time-optimal performance, has become a benchmark for evaluating agile trajectory generation and robust control strategies under extremely high-speed conditions.
Despite such progress, achieving time-optimal high-speed flight in real time, particularly in unknown environments, remains a formidable challenge. Specifically, the limited onboard resources of MAVs demand 
efficient system architectures and algorithms 
capable of handling complex dynamics models, unexpected disturbances and safety constraints in real time. 

\begin{figure}[t!]
    \vspace{-2pt}  
    \centering
    \includegraphics[width=\columnwidth, clip]{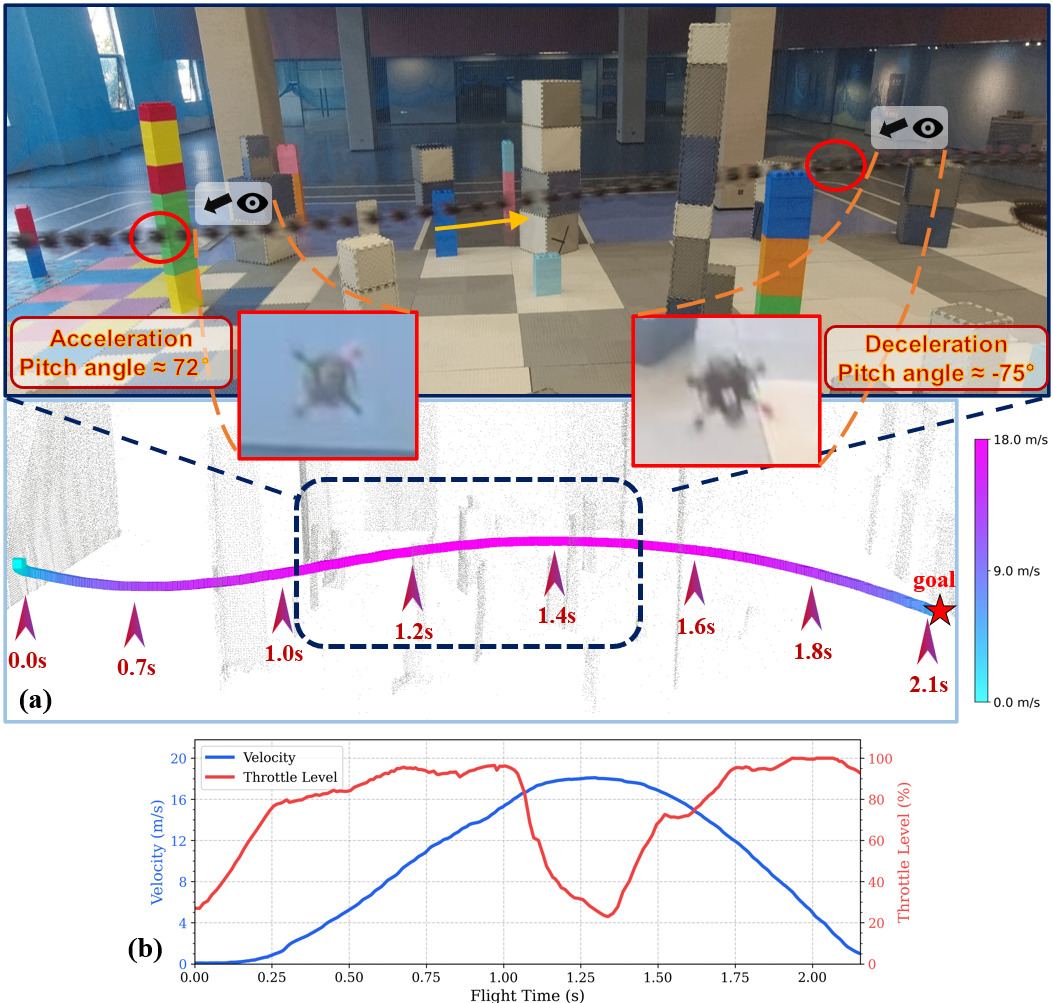}
    \caption{ 
    (a) High-speed autonomous flight trajectory of our method in the real world and point-cloud map visualization. 
    (b) The corresponding throttle and velocity profiles. 
    The MAV completes a 20\,m flight in 2.1\,s and reaches a peak speed of 18.1\,m/s in 1.2\,s.  
    Side and forward views both reveal that the MAV attains an aggressive pitch angle of approximately 
    $75^\circ$ during acceleration and deceleration.
    }
    \vspace{-8pt}  
    \label{fig:real_highspeed}
\end{figure}

Several works have focused on achieving online autonomous flight for MAV systems in unknown and cluttered environments. 
In early efforts, Richter et al.~\cite{richter2016polynomial} achieve aggressive flights in known environments by optimizing polynomial trajectories based on differential flatness~\cite{mellinger2011minimum, faessler2017differential} of the quadrotor to improve 
systematic 
efficiency. 
For real-time collision avoidance in unknown environments, a complete MAV system~\cite{oleynikova2016continuous} is introduced based on a local polynomial trajectory planner.
In \cite{mohta2018fast}, a MAV navigates through cluttered environments in a receding-horizon fashion, where only a local map is maintained for the local planner.
In this fashion, an adequate replanning frequency is required, since high computation delays severely limit both flight speed and success rate~\cite{ren2025safety}.
To enable efficient replanning,
Fast-Planner~\cite{zhou2019robust} employs B-spline for path-guided optimization, while EGO-Planner-v2~\cite{zhou2022swarm} leverages control effort minimizer (MINCO)~\cite{wang2022geometrically} for spatial-temporal optimization. 
However, the aforementioned methods cannot systematically enable high-speed flight. 
One reason for such low-speed flight is that visual perception can not provide a sufficiently long detection range (typically below 5\,m) and reaction time for avoiding obstacles at high speeds. A straightforward solution is to employ LiDAR sensors, which deliver a sufficiently wide sensing range.
For instance, Ren et al.~\cite{ren2022bubble} adopt a LiDAR-based receding horizon planning to enable high-speed flight in the wild.
More recently, the state-of-the-art method SUPER~\cite{ren2025safety} demonstrates LiDAR-based safety-assured high-speed navigation in unknown environments using two-trajectory planning,
with trajectory tracking by on-manifold MPC~\cite{lu2023On}.
Even so, these methods do not address the other reason for conservative flight, namely that the separation of planning and control into a multi-stage pipeline increases system latency.
To address this, integrated planning and control (IPC) methods employ model predictive control (MPC) with a point-mass model~\cite{liu2023integrated} or collective-thrust-input model~\cite{yuan2025safety} for agile obstacle avoidance.
Another class of approaches is end-to-end learning frameworks, which inherently 
adopt an integrated pipeline.
Loquercio et al.~\cite{loquercio2021learning} apply imitation learning to enable high-speed flight in the wild
with inputs of depth images and partial states. 
Zhang et al.~\cite{zhang2024back} exploit a differentiable physics engine with a point-mass model 
and depth rendering to learn agile flight across diverse scenarios.  
Nevertheless,
a substantial gap remains compared with time-optimal flights.
First,
when constraints exist, optimal solutions cannot in general be represented by polynomial splines\cite{wang2022geometrically}, as used in most optimization-based frameworks. 
Second, 
mismatches between state-level constraints (on velocity/acceleration) used in the above frameworks and actual input constraints in real systems can cause tracking failure or excessive conservatism.
Finally,
all of the aforementioned frameworks rely on simplified dynamics rather than full dynamics, and employ conservative time allocations to satisfy real-time computational limits onboard MAVs.

In terms of time-optimal planning and control,
several studies of drone racing have focused on minimizing flight time by fully exploiting the system dynamics, including the low-level actuator inputs (e.g., single-motor thrust),
under the given MAV platform and maximum thrust-to-weight ratio (TWR).
Remarkably,
Foehn et al. optimize global trajectories through waypoints using a complementary progress constraint to achieve time-optimal planning~\cite{foehn2021time}.
However, such methods are computationally expensive and unsuitable for online replanning.
To address this issue, reinforcement learning (RL)
and model predictive contouring control (MPCC)
have been introduced.
Song et al. use an RL controller to reach a peak speed of 30 m/s with a maximum TWR of 12, with external computation and sensing~\cite{song2023reaching}.
Kaufmann et al. similarly fly the MAV at its physical limits, using the onboard camera~\cite{kaufmann2023champion}.
These RL methods can learn agile flight policies, yet often suffer from limited generalization to unseen scenarios 
and require offline training.
Time-optimal MPCC provides an integrated planning-control framework that optimizes progress along a reference path while effectively handling
dynamics and disturbances, achieving a peak speed over 16\,m/s with a maximum TWR of 3.3~\cite{romero2022model}. 
Romero et al. propose a sample-based method to generate a point-mass model reference for MPCC~\cite{romero2022time}.
Nevertheless, 
all these works assume a known environment 
(e.g., a racing track).
Therefore,
their integration into autonomous navigation systems and generalization to safely traverse unknown environments, especially under limited onboard resources, remain unclear.

To handle obstacle avoidance, the distance and gradient information to obstacles, derived from Euclidean signed distance field (ESDF)~\cite{oleynikova2017voxblox} or from projected gradient information~\cite{zhou2022swarm}, are commonly integrated into gradient-based trajectory optimization as penalty terms.
However, such gradient-based methods struggle to guarantee safety in high-speed flight, since they do not provide explicit safety boundaries for optimization.
When obstacles are suddenly detected during high-speed flight, 
soft penalties may cause the trajectory optimization to violate constraints, potentially leading to collisions with obstacles.
One class of solutions for solving this problem employs control barrier function (CBF) constraints to ensure the safety of dynamic systems by defining safe sets for quadrotor states~\cite{yuan2025safety}.
However, while CBF provides formal safety guarantees, it can be sensitive to input limits and constraint conflicts, and often requires careful tuning to ensure feasibility and practical effectiveness.
Another widely adopted approach abstracts obstacle-free spaces as convex polytopes~\cite{wang2025fast}, also referred to as a safe flight corridor (SFC)~\cite{liu2017planning}.
This compact, convex representation significantly simplifies optimization and enhances planning efficiency and reliability in cluttered environments.
Nonetheless, while the aforementioned works can realize safe obstacle avoidance across diverse scenarios, their planning-control frameworks, algorithm implementations, and system integrations
fail to jointly achieve the time-optimal motion performance in unknown environments.



Despite such advances, 
existing works
do not concurrently satisfy the requirements of
time-optimal, high-speed autonomous flight with safe obstacle avoidance in unknown environments for MAVs. 
In this paper,
we propose a learning-accelerated online time-optimal integrated planning and control framework (LOONG),
which builds upon our previous work in drone racing~\cite{guan2025learning} and is inspired by SUPER\cite{ren2025safety}.
The key contributions of this work are as follows:

\begin{enumerate}[1)]

    \item A time-optimal integrated planning and control framework (Fig.~\ref{fig:system}) for MAVs is proposed to achieve autonomous flight in unknown, cluttered environments. 
    This framework is characterized by high-frequency replanning (100\,Hz).
    At each replanning step, time-optimal polynomial reference generation is accelerated through learning-based time allocation. 
    Following this reference, the planning and control are solved simultaneously in one optimization problem to further enhance system performance.

\begin{figure}[t!] 
    \centering
    \includegraphics[width=0.49\textwidth]{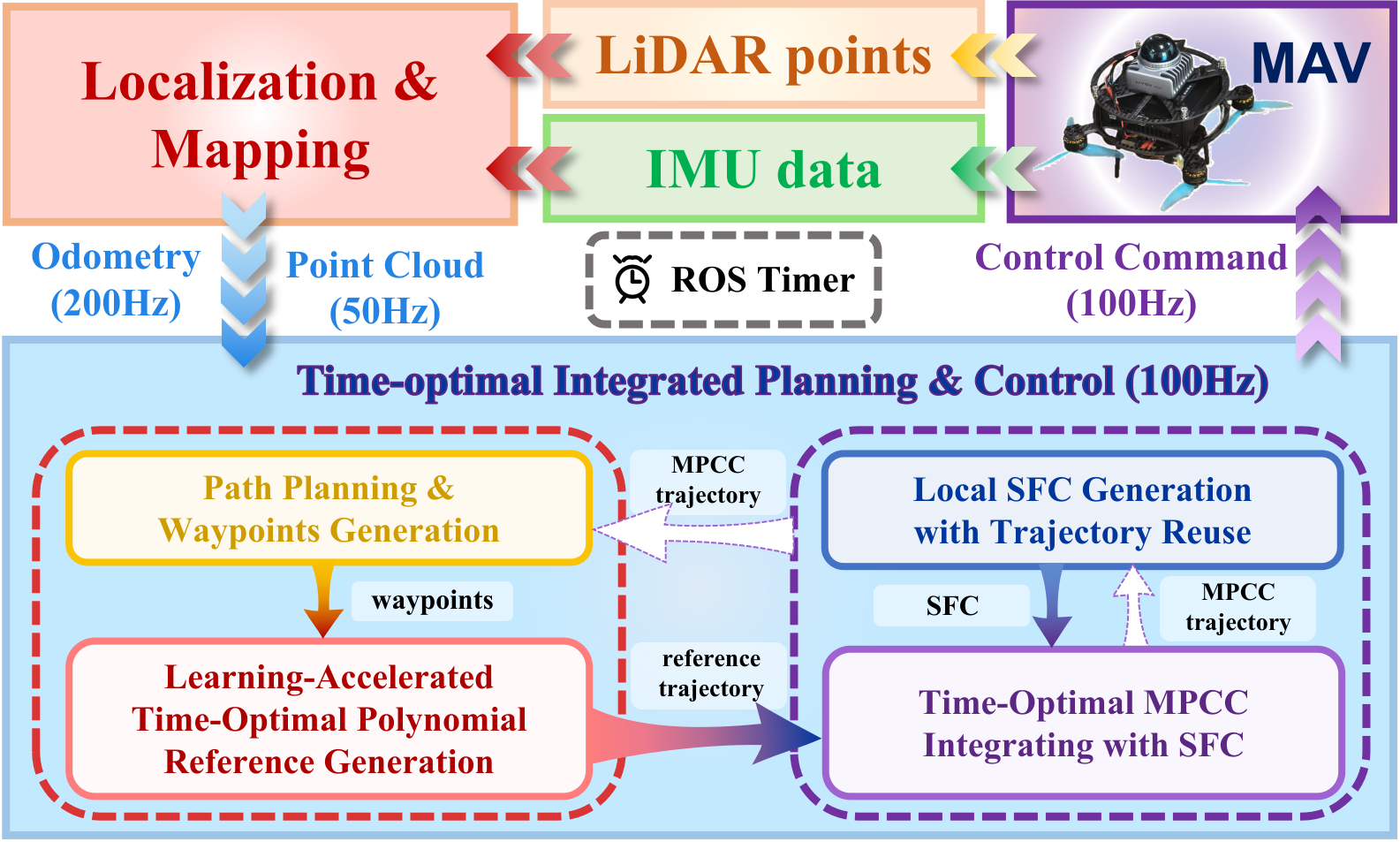} 
    \caption{
    LOONG is an integrated planning-control framework 
    designed for time-optimal autonomous flight. 
    By leveraging odometry and a point-cloud map generated from LiDAR and IMU sensors, LOONG directly outputs angular velocity and throttle commands to the flight control unit. It comprises a computationally lightweight reference generation module as the frontend, and a high-performance optimal control module as the backend. 
    Online replanning is guided by the previous MPCC trajectory to achieve both aggressive and safe flight.
    }

    \label{fig:system}
    \vspace{-8pt}
\end{figure}

    \item 
    An efficient MPC is proposed that integrates time-optimal MPCC with SFC for MAVs integrated planning and control, especially in high-speed flight,
    leveraging the MAV's full dynamics.
    By imposing SFC constraints at variable horizon steps, 
    we concurrently satisfy both aggressiveness and safety requirements.
    The outputs of MPCC are directly used as actual execution commands.



    

    \item 
    We integrate the proposed method into a 
    LiDAR-based
    fully autonomous quadrotor system
    and extensively evaluated the system in both simulation and real-world environments.
    In simulation, our framework outperformed the state of the art for 
    autonomous navigation in terms of aggressiveness while achieving obstacle avoidance.
    In the real world, the MAV achieved 
    a peak speed of 18\,m/s in a 20\,m flight 
    and succeeded over 10 consecutive trials from different start points.

\end{enumerate}

\section{Learning-Accelerated Online Time-Optimal Reference Trajectory Generation}
\label{sec:ref_traj_gen}

The system 
architecture 
based on the proposed 
planning and control
framework is illustrated in Fig.~\ref{fig:system}. 
It operates upon 
a point-cloud map and
characterized by a learning-accelerated time-optimal trajectory generator (Section\,\ref{sec:ref_traj_gen}) with a processing latency below 1\,ms to swiftly respond to environmental changes during high-speed flight, and a 
safe-enhanced 
time-optimal 
MPCC
algorithm (Section\,\ref{sec:mpcc_overall}), which runs at 100\,Hz.


In time-optimal autonomous flight, efficient trajectory planning is crucial for ensuring safety at high speeds due to the limited perception range. 
This section employs an efficient and safe front-end path planning method, along with a learning-based time-allocation approach, for reference trajectory generation. This method significantly enhances the efficiency of generating time-optimal polynomial trajectories to serve as exploratory references for MPCC, which requires a good approximation of the time-optimal policy to enhance the stability and quality of the solution.
\vspace{-2pt}\subsection{Path Planning and Waypoints Generation}
\label{sec:front-end}

\begin{figure}[t!] 
    \centering
    \includegraphics[width=0.49\textwidth]{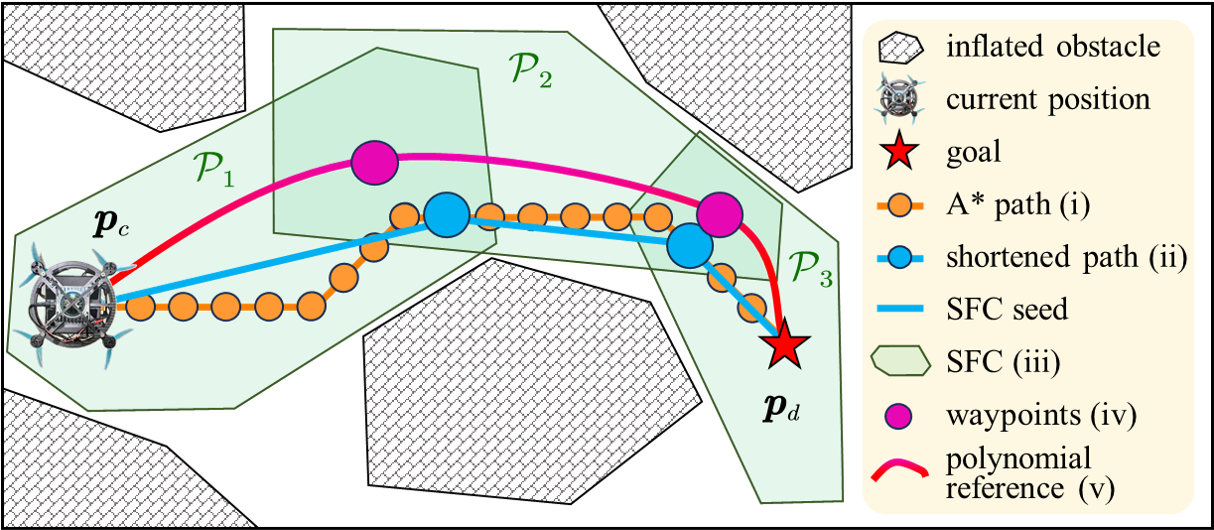} 
    \caption{Path planning and waypoints generation process involves (i) an A*-based path search, (ii) shorten and line seed generation, (iii) SFC convex decomposition (\rmnum{4}) and computing overlapping centers as waypoints for (\rmnum{5}) subsequent trajectory generation.}
    \label{fig:pipeline_a}
    \vspace{-8pt}
\end{figure}

The front-end path planning aims to efficiently generate a safe global path
including discrete positions or waypoints
for obstacle avoidance
from the current position $\bm{p}_c$ to the goal position $\bm{p}_d$, as depicted in Fig.~\ref{fig:pipeline_a}. 
Initially, an A* path (\rmnum{1}, orange points) search algorithm
is employed to find a collision-free path connecting $\bm{p}_c$ and $\bm{p}_d$.
To find the shorten path (\rmnum{2}), the 
discretized 
A* path is then refined
by iteratively finding the furthest visible point from the end of the last line segment (blue lines).

Since the shortened path typically lies near obstacle boundaries, inspired by~\cite{ren2025safety}, 
we want to enhance safety by keeping waypoints away from obstacles, as this improves onboard sensor visibility and enlarges the free space available for local trajectory planning. To this end, we introduce the concept of a safe flight corridor (SFC), which consists of a set of convex polytopes derived from seeds (e.g., point or line) that explicitly exclude obstacles~\cite{wang2025fast}.
The polytope~$\mathcal{P}$ is defined as the intersection of $m$ half-spaces:
\begin{equation}
\mathcal{P} = \{ \bm{y} \in \mathbb{R}^n \mid \bm{A} \bm{y} \leq \bm{b} \}, \quad \bm{A} \in \mathbb{R}^{m \times n}, \quad \bm{b} \in \mathbb{R}^m
\label{eq:polytope}
\end{equation}
where $\bm{y}$ are the states of the MAV, $m$ is the number of hyperplanes of a convex polytope, $\bm{A} \in \mathbb{R}^{m \times n}$ representing the normal vectors of hyperplanes, and $\bm{b} \in \mathbb{R}^m$ representing the constants of hyperplanes. 
Here, we set $n=3$ to impose SFC constraints on position space.

Therefore, as shown in Fig.~\ref{fig:pipeline_a}, each line segment derived from the shortened path (\rmnum{2}) can serve as a seed for SFC convex decomposition (\rmnum{3}, green polytopes).
Then, the centers of the overlapping regions between adjacent polytopes are selected as waypoints (\rmnum{4}, purple points) for subsequently generating time-optimal polynomial reference trajectories (\rmnum{5}).
Specifically,
to enhance efficiency and ensure compatibility with LiDAR point clouds, we adopt the configuration-space iterative regional inflation (CIRI) algorithm~\cite{ren2025safety}, which directly operates on point clouds to extract convex polytopes.



\vspace{-2pt}\subsection{Time-Optimal Polynomial Trajectory Optimization}
\label{sec:ref_poly_traj_gen}


After 
path planning, a sequence of next \( M \) waypoints \( \bm{p}_i \), \( \forall i \in \{1, \dots, M\} \) is obtained.
To enable high-speed flight, 
This section focuses on generating a time-optimal reference under the finite-degree polynomial representations passing through all \( \bm{p}_i \).
Typically, 
when considering the full quadrotor dynamics,
the nonlinearity 
of constraints
make model-based time-optimal trajectory optimization computationally demanding. By exploiting the differential flatness~\cite{mellinger2011minimum, faessler2017differential},
the quadrotor's trajectory can be expressed through its flat outputs:
$\bm{\sigma} = [p_x, p_y, p_z, \psi]^T \in \mathbb{R}^3 \times S^1$,
where \( \bm{p}=[p_x, p_y, p_z]^T \in \mathbb{R}^3 \) denotes the position, and \( \psi \in S^1 \) indicates the yaw angle.
As such, 
constraints can be applied discretely to the linear combination of the 0-th to \(s\)-th order derivatives of \( \bm{\sigma} \). To extract state information at any 
timestamp, 
a common method involves parameterizing the trajectory as a \( M \)-piece polynomial 
\( \bm{\sigma}(t) \), 
with each piece represented by a \( N = 2s - 1 \) degree polynomial
$\bm{\sigma}_i(t) = \mathbf{c}_i^\top \bm{\beta}(t - t_{i-1}), t \in [t_{i-1}, t_i],$where $\bm{\beta}(t) = [1, t, t^2, \dots, t^N]^\top$.

The polynomial trajectory 
\( \bm{\sigma}(t) \) is encoded through its coefficient matrix $\mathbf{c}$ and time allocation intervals $\mathbf{T}$:
\begin{equation}
\begin{aligned}
\mathbf{c} &= [\mathbf{c}_1^\top, \dots, \mathbf{c}_M^\top]^\top \in \mathbb{R}^{2M s \times 4} \\
\mathbf{T} &= [T_1, \dots, T_M]^\top \in \mathbb{R}^{M}_{>0}
\end{aligned}
\end{equation} 
where \( T_i \) represents the duration of the \( i \)-th segment. The timestamp for each waypoint is given by $t_i = \sum_{j=1}^{i} T_j$ 
and the total duration of the trajectory is \( T = \|\mathbf{T}\|_1 \).

Using the minimal-control polynomial trajectory~\cite{wang2022geometrically}, 
the 
coefficient matrix is given by:
\begin{equation}
\label{eq:linear_system}
\textbf{M(T)}\mathbf{c} = \mathbf{b}
\end{equation}
where \( \textbf{M} \in \mathbb{R}^{2Ms \times 2Ms} \) is a nonsingular banded matrix determined solely by the time allocation \( \mathbf{T} \), and \( \mathbf{b} \in \mathbb{R}^{2Ms \times 4} \) contains the specified derivatives (i.e., the positions and their respective derivatives) at the boundaries of $\bm{\sigma}_i(t)$.

The unique solution \( \mathbf{c} \) can be obtained by solving 
(\ref{eq:linear_system})
given $\mathbf{T}$.
Within the polynomial representation, the time-optimal trajectory generation problem 
is defined as:
\begin{equation}
\begin{alignedat}{2}
    & \!\!\!\min_{\mathbf{T}} \quad && T, \\
    & \text{s.t.} \quad && \mathcal{F}(\bm{\sigma}(t), \ldots, \bm{\sigma}^{(s)}(t)) \preceq 0, \quad \forall t \in [0, T], \\
    & && \bm{\sigma}^{[s-1]}(0) = {\bm{\sigma}}_o, \quad \bm{\sigma}^{[s-1]}(T) = {\bm{\sigma}}_f, \\
    & && \bm{\sigma}(t_i) = {\bm{\sigma}}_i, \quad 1 \leq i \leq M-1
\end{alignedat}
\label{eq:optimization_problem}
\end{equation}
where \( {\bm{\sigma}}_o \in \mathbb{R}^{4 \times s} \) denotes the initial state, and \( {\bm{\sigma}}_f \in \mathbb{R}^{4 \times s} \) represents the final state. \( {\bm{\sigma}}_i \) represents the combination of \( \bm{p}_i \) and a fixed yaw angle, and \( \mathcal{F} \) represents the dynamically feasible constraints imposed on the system.

In this paper,
we represent the trajectory using 5-degree minimum-jerk polynomials, with \( s = 3 \) to satisfy the collective thrust and bodyrates constraints.
The optimization problem (\ref{eq:optimization_problem}) can be solved in two layers, with the outer layer optimizing the duration $T_i$ and the inner layer solving (\ref{eq:linear_system}) for a given $T_i$ to compute the coefficients $\mathbf{c}_i$ of each polynomial segment.
However, the outer 
layer optimization
problem is non-convex, and solving it with
nonlinear 
constraints 
to find feasible solutions
can be computationally intensive. 
Furthermore, the quality and success rate of the optimization are sensitive to the initial guess of $\mathbf{T}$, which makes it difficult to stably generate valid trajectories.
Therefore, a stable and efficient method is required to optimize time allocation and generate references for MPCC during high-frequency planning in flight.

\vspace{-2pt}\subsection{Accelerating Time Allocation by Imitation Learning}
\label{sec:NN_acc}
As stated above,
the main challenge in generating a time-optimal 
trajectory
lies in solving the computationally intensive minimum-time problem (\ref{eq:optimization_problem}), primarily due to the complexity of time allocation and the sensitivity to initial conditions. Fortunately, once the optimal time allocation $\mathbf{T}^*$ is provided, 
the polynomial trajectory can be generated by solving (\ref{eq:linear_system}) with linear complexity, requiring minimal computational resources.

\begin{figure}[t!] 
    \centering
    \includegraphics[width=0.49\textwidth]{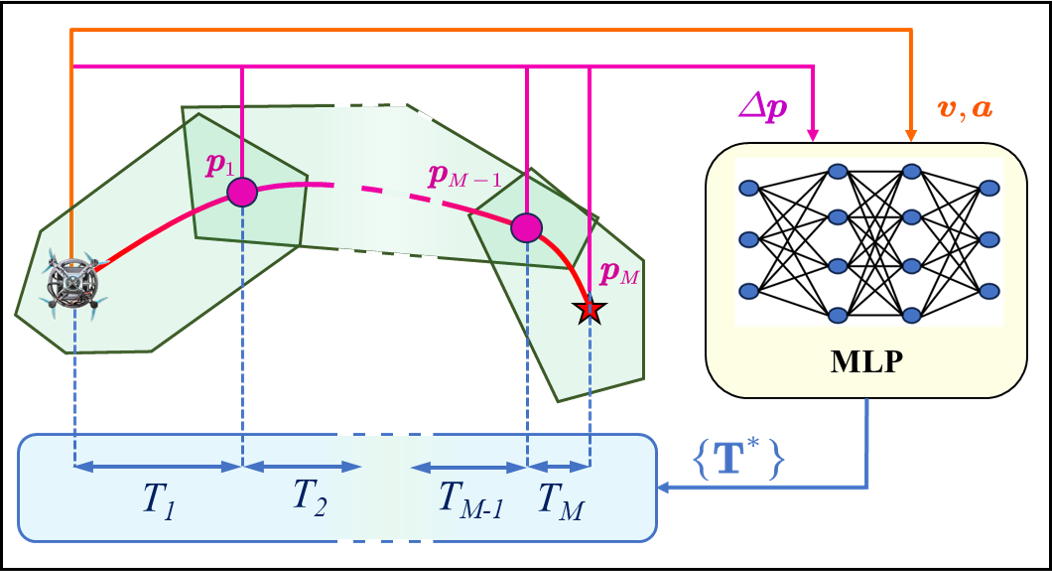} 
    \caption{Neural network inferences time allocation from current states and waypoints to efficiently generate a multi-piece polynomial as reference trajectory.}
    \label{fig:pipeline_b}
    \vspace{-8pt}
\end{figure}

To accelerate the time allocation process in real time, we utilize a lightweight neural network that mimics the time allocation of expert trajectories, as shown in Fig.~\ref{fig:pipeline_b}. The network consists of a multi-layer perceptron (MLP) with two hidden layers, each containing 512 neurons. 
The MLP's input is a one-dimensional vector comprising the MAV's current velocity $\bm{v} \in \mathbb{R}^3$ and acceleration $\bm{a} \in \mathbb{R}^3$, along with the relative positions to the next $M$ waypoints $\Delta \bm{p} \in \mathbb{R}^{3 \times M}$.
The output is the sequence of optimal time allocations $\mathbf{T}^* \in \mathbb{R}^M$.
Note that during flight in unstructured environments, the number of waypoints obtained by the front-end path planner may vary. 
To address this, we set a sufficiently large $M$ to maintain the neural network architecture and pad the input and output with placeholder values to preserve the expected dimensions.
Subsequently, the 
vector $\mathbf{T}^*$ is utilized 
alongside 
$\bm{p}_i $
to solve 
(\ref{eq:linear_system}), yielding the polynomial coefficients $\mathbf{c}$ and thus the optimal polynomial trajectory.
For dataset generation, 
We use MINCO \cite{wang2022geometrically} to connect the current states and the sampled relative waypoints
$\bm{p}_i$ 
via minimum-control polynomials. The waypoints $\bm{p}_i$ and their corresponding durations $T_i$ are extracted to form the dataset used to train the neural networks. 
Readers are referred to our previous work \cite{guan2025learning} for details of the time-allocation network 
applied to
time-optimal planning.



In summary, 
this method can generate a dynamically feasible and near time-optimal trajectory in real time under polynomial representations.
However, polynomial trajectory optimization is a relaxation of the time-optimal control problem for MAVs navigating through multiple waypoints, which inevitably leaves a non-negligible gap compared to true time-optimal navigation.
This 
gap
stems primarily from the inherent smoothness of finite-degree polynomials, which cannot fully exploit
control inputs\cite{foehn2021time}.

\section{Time-Optimal Model Predictive Contouring Control with Collision Avoidance}
\label{sec:mpcc_overall}

This section presents a control method, referred to as MPCC, that tracks the reference trajectory in a truly time-optimal manner while accounting for the full quadrotor dynamics and obstacle avoidance.



\vspace{-2pt}\subsection{Quadrotor Dynamics} 
\label{sec:dynamics}

The quadrotor's state space is expressed as \( \bm{x} = 
\begin{bmatrix}
\bm{p}, \bm{q}, \bm{v}, \bm{\omega}
\end{bmatrix}^T \),
where \( \bm{p} \in \mathbb{R}^3 \) represents the position, \( \bm{q} \in \mathbb{SO}(3) \) denotes the attitude quaternion, \( \bm{v} \in \mathbb{R}^3 \) is the linear velocity, and \( \bm{\omega} \in \mathbb{R}^3 \) represents the bodyrates in the body frame. 
The system inputs include the collective thrust \( \bm{F}\) and body torques \( \bm{\tau} \).
The dynamic equations 
are expressed as:

\begin{equation}
\label{eq:dynamics}
\begin{aligned}
\dot{\bm{p}} &= \bm{v}, \\
\dot{\bm{q}} &= \frac{1}{2} \bm{q} \odot 
\begin{bmatrix}
0 \\ \bm{\omega}
\end{bmatrix}, \\
\dot{\bm{v}} &= \bm{g} + \frac{1}{m} \mathbf{R}(q) \bm{F} - \mathbf{R}(\bm{q}) \bm{D} \mathbf{R}^T(\bm{q}) \cdot \bm{v}, \\
\dot{\bm{\omega}} &= \mathbf{J}^{-1} \left( \bm{\tau} - \bm{\omega} \times \mathbf{J} \bm{\omega} \right)
\end{aligned}
\end{equation}
where \( \odot \) represents Hamilton quaternion multiplication, \( \mathbf{R}(\bm{q}) \) is the quaternion rotation matrix, \( m \) is the quadrotor mass, \( \mathbf{J} \) is the inertia matrix, and \( \bm{D} = \text{diag}(d_x, d_y, d_z) \) represents the drag coefficients~\cite{faessler2017differential}.

Additionally, the input of \( \bm{F} \) and \( \bm{\tau} \) is decomposed into individual rotor thrusts $\bm{f}=[f_1,f_2,f_3,f_4]$:

\begin{equation}
\bm{F} = 
\begin{bmatrix}
0 \\ 
0 \\ 
\sum f_i
\end{bmatrix}, 
\quad \text{and} \quad 
\bm{\tau} = 
\begin{bmatrix}
\frac{l}{\sqrt{2}} (f_1 + f_2 - f_3 - f_4) \\
\frac{l}{\sqrt{2}} (-f_1 + f_2 + f_3 - f_4) \\
c_{\tau}(f_1 - f_2 + f_3 - f_4)
\end{bmatrix}
\end{equation}
where $f_i$ is the thrust at rotor $i\in \{1,2,3,4\}$, \( l \) represents the quadrotor's arm length and \( c_{\tau} \) is the rotor's torque constant.

\vspace{-2pt}\subsection{Time-Optimal MPCC Integrating with SFC}
\label{sec:mpcc_with_sfc}

\begin{figure}[t!] 
    \centering
    \includegraphics[width=0.49\textwidth]{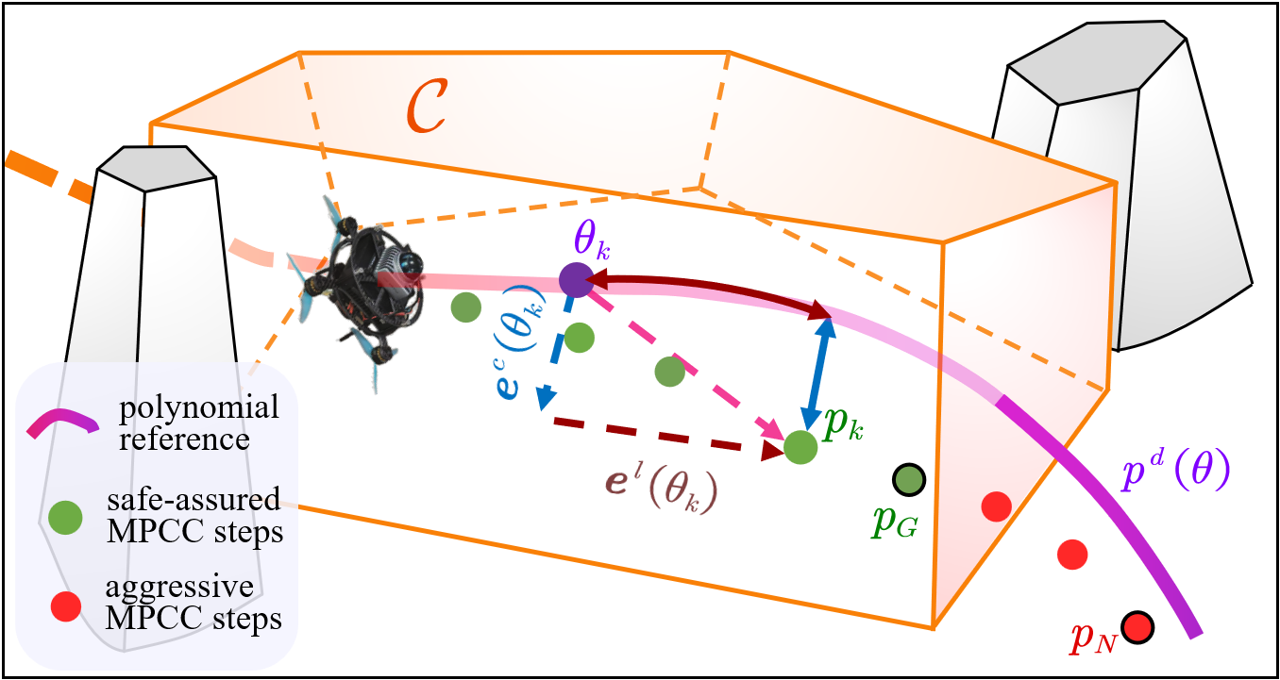} 
    \caption{3D MPCC diagram with local SFC constraints 
    applied to the first $G$ steps 
    to enable 
    time-optimal
    flight while avoiding obstacles.}
    \label{fig:pipeline_d}
    \vspace{-5pt}
\end{figure}

In contrast to solving trajectory planning and control separately, model predictive contouring control (MPCC) 
allows robust tracking of a continuously differentiable path while maximizing 
the traverse speed
in a receding horizon fashion by solving the time allocation and control problems simultaneously in real time~\cite{romero2022model}.  
In time-critical scenarios, a time-optimal reference path (e.g., time-optimal point-mass model trajectory or time-optimal polynomial trajectory~\cite{guan2025learning}) helps the MPCC find a good approximation of the time-optimal policy.
For the MPCC, deviations from the reference path are permitted to enhance aggressiveness. 
Nevertheless, 
in cluttered environments,
sacrificing tracking accuracy 
for time-optimal performance 
can result in unsafe flight trajectories. 
Thus, to ensure safety, this section proposes a 
variant MPCC 
to constrain a certain horizon of the predicted trajectory 
within the known free space based on SFC, 
enabling agile obstacle avoidance 
at a high replanning frequency,
as shown in Fig.~\ref{fig:pipeline_d}. 

The contouring control problem transforms the trajectory to be tracked, parameterized by time, into an arc-length parameterized trajectory. 
The arc length of the reference path (or progress) is denoted as $\theta$, with the arc length at timestep $k$ denoted as $\theta_k$.  
To incorporate the dynamics (\ref{eq:dynamics}) within the MPCC formulation, the state space is augmented with progress dynamics by adding $\theta$ and progress speed $v_{\theta}$ as virtual states: 
\begin{equation}
\bm{\bar{x}} = [\bm{p}, \bm{q}, \bm{v}, \bm{w}, \bm{f}, \theta, v_{\theta}]^T, 
\quad \bm{\bar{u}} = [\Delta \bm{f}, \Delta v_{\theta}]^T
\label{eq:state_input_augmented}
\end{equation}
where the progress acceleration
$\Delta v_{\theta}$
is introduced as a virtual input, following typical contouring controller implementations \cite{romero2022model},
to avoid uncontrolled rapid change of $v_{\theta}$, which would result in noisy control inputs. The dynamics of the augmented states are:
\begin{equation}
\dot{\bm{f}} = \Delta \bm{f}, \quad \dot{\theta} = v_{\theta}, \quad \dot{v}_{\theta} = \Delta v_{\theta}\\
\label{eq:dynamics_augmented}
\end{equation}

The MPCC problem is formulated as follows :
\begin{equation}
\label{eq:MPCC}
\begin{aligned}
\pi(\bm{x}) = \underset{\bm{\bar{u}}}{\mathrm{argmin}} &\sum_{k=0}^{N} \big(  \| \bm{e}^l(\theta_k) \|^2_{q_l} + \| \bm{e}^c(\theta_k) \|^2_{q_c} \\
+ \| \bm{\omega}_k \|^2_{{Q_\omega}} &+ \| \Delta \bm{f}_k \|^2_{{R}_{\Delta \bm{f}}} + \| \Delta v_{\theta_k} \|^2_{R_{\Delta v}} - \mu v_{\theta, k} \big)  \\
\text{subject to} \quad & \bm{x}_0 = \bm{\bar{x}} \\
& \bm{\bar{x}}_{k+1} = \bm{\bar{x}}_k + d_t \cdot f(\bm{\bar{x}}_k, \bm{\bar{u}}_k)  \\
& \bm{\omega}_{min} \leq \bm{\omega} \leq \bm{\omega}_{max}  \\
& \bm{f}_{min} \leq \bm{f} \leq\bm{f}_{max}  \\
& 0 \leq v_\theta \leq {v}_{\theta,max}  \\
& \Delta \bm{f}_{min} \leq \Delta \bm{f} \leq \Delta \bm{f}_{max} \\
& \Delta v_{\theta,min} \leq \Delta v_\theta \leq \Delta v_{\theta,max}  \\
\end{aligned}
\end{equation}
where the contour error $\bm{e}^c(\theta_k)$ and lag error $\bm{e}^l(\theta_k)$ represent the perpendicular and tangential errors between the position $p_k$ and its projection on the reference path $p^d(\theta)$ at timestep $k$, as in Fig.~\ref{fig:pipeline_d}. The $q_c$, $q_l$, and $\mu$ are the weights for the contour error, lag error, and progress term, respectively. $N$ represents the prediction horizon, and
$f(\bm{\bar{x}}_k, \bm{\bar{u}}_k)$
denotes the dynamics derived by discretizing (\ref{eq:dynamics}) and (\ref{eq:dynamics_augmented}) with time step $d_t$.



Although the waypoints generated in Section\,\ref{sec:front-end} are guaranteed to be obstacle-free, the resulting polynomial reference trajectories may still collide with obstacles, since no constraints are imposed on the trajectory segments between the waypoints.
Moreover, due to the inherent trade-off between tracking accuracy and time optimality in MPCC, the resulting trajectory may deviate from the reference even when the reference itself is collision-free, potentially leading to collisions with obstacles.
Furthermore, (\ref{eq:polytope}) optimistically assumes that unseen regions are obstacle-free.
Thus,
to ensure safety, 
the planned trajectory of the MPCC must be rigorously constrained to remain within the known free space.
From point cloud data under partial environmental knowledge, 
it is guaranteed
that a polytope $\mathcal{C}$ extracted by CIRI remains entirely within the known free space, provided that the input point cloud forms a sufficiently dense depth image from LiDAR sensors and that the input seed includes the current MAV position~\cite{ren2025safety}. 
Hence, similar to (\ref{eq:polytope}), a local SFC $\mathcal{C}$ is qualified as a strict constraint for position $\bm{p}$ to ensure safety:
\begin{equation}
\mathcal{C} = \{ \bm{p} \in \mathbb{R}^3 \mid \bm{A} \bm{p} \leq \bm{b} \}, \quad \bm{A} \in \mathbb{R}^{m \times 3}, \quad \bm{b} \in \mathbb{R}^m
\end{equation}

For each planning cycle,
the naive idea is to
directly incorporate the local SFC constraints $\mathcal{C}$ into the 
MPCC predicted positions over the entire future horizon
$N$.
However,
when unknown obstacles are detected during flight, environmental changes can cause frequent shrinking and expanding of the local SFC, which is particularly true during high-speed flight or in unstructured, dense environments. 
%
If SFC constraints are applied to the entire horizon, a suddenly narrowed SFC may result in its range being smaller than the MPCC's trajectory length. 
Thus, it may decrease global flight speed, leading to conservative performance. 
Moreover, as a well-known drawback of optimization-based controllers, imposing excessively tight constraints may undermine the solver's stability and success rate~\cite{KOHLER2024100929}.

Considering the high replanning frequency and efficient time allocation of the proposed method, 
since MPC follows a receding-horizon scheme and applies only the first control input at each iteration, constraining the immediate state should guarantee that the quadrotors remain within the safe region. 
However, for real-time implementation, the MPCC employs an iterative scheme for computational efficiency, 
which requires multiple rolling iterations to obtain a feasible solution~\cite{verschueren2022acados}.
During runtime, constraining only the next-step state leads to suboptimal obstacle avoidance performance and may compromise safety.
To strike a balance between aggressiveness and safety, we define $G<N$ as the 
safe
horizon and only impose constraints on the first $G$ steps of the MPCC:
\begin{equation}
\bm{A} \bm{p}_k \leq \bm{b}, \quad k \in \{1, \ldots, G\}
\label{eq:Gsfc_constraint}
\end{equation}
where $\bm{p}_k$ is the predicted position computed by MPCC at timestep $k$.
By embedding (\ref{eq:Gsfc_constraint}) into (\ref{eq:MPCC}), aggressive yet safe flight can be achieved,
as depicted in Fig.~\ref{fig:pipeline_d}. 





\begin{figure}[t!] 
    \centering
    \includegraphics[width=0.49\textwidth]{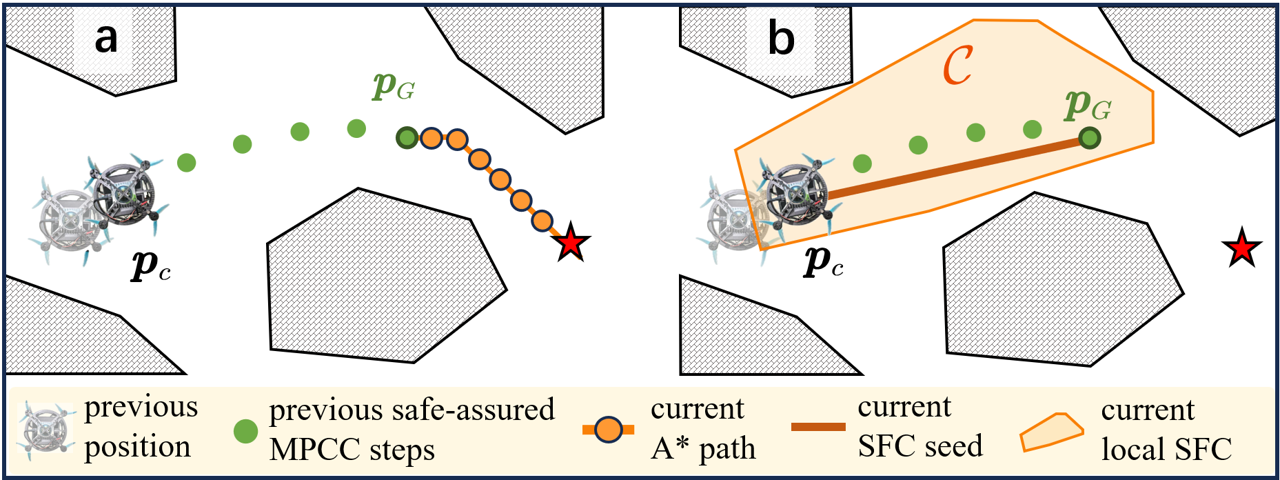} 
    \caption{
     Illustration of the strategy for maintaining consistency across two consecutive planning loops. (a) The A* algorithm is initialized from $p_G$, the $G$-th state of the trajectory generated in the previous planning loop. (b) The line segment connecting the current state $p_c$ and $p_G$ is used as the seed for generating the SFC.
     }
    \label{fig:pipeline_c}
    \vspace{-6pt}
\end{figure}

For high-speed flight in partially known environments, 
the continuously updating map necessitates high-frequency replanning for obstacle avoidance.
However,
When the local SFC $\mathcal{C}$ and front-end path are generated,
the detection of previously unseen obstacles can lead to inconsistencies in path topology between consecutive planning loops, which may result in suboptimal or even unstable behavior of the MPC-based scheme.
To address this issue, we propose a replanning method that ensures a smooth and consistent transition between trajectories generated in consecutive planning loops, which is called the \textbf{\textit{trajectory reuse}} strategy.
At current planning loop, the A* algorithm is initialized from $p_G$, which corresponds to the $G$-th state of the trajectory generated in the previous planning loop, while the states between the current state $p_c$ and $p_G$ are guaranteed to be safe by the SFC constructed in the previous loop (Fig.~\ref{fig:pipeline_c}(a)).  Moreover, since both $p_c$ and $p_G$ lie within the convex polytope of the previous SFC, which is obstacle-free by construction, they can be directly used as valid seeds for convex decomposition to generate the SFC $\mathcal{C}$ for the current planning loop (Fig.~\ref{fig:pipeline_c}(b)). This strategy ensures the consistency of the MPCC formulation across consecutive planning loops.
The effectiveness of trajectory 
reuse
is demonstrated via ablation studies in Section\,\ref{sec:exp}. 

\section{Experiments and Analysis}
\label{sec:exp}


\vspace{-2pt}\subsection{Simulation Experiments}
\subsubsection{Experimental Setup} 
The proposed approach is evaluated extensively through both simulation and real-world experiments.
The basic parameters
are kept consistent in both simulation and real-world experiments, as shown in Table \ref{tab:param}. 
The MPCC is deployed on acados \cite{verschueren2022acados}, using SQP-RTI for real-time computation, with the control loop running at 100 Hz. We use a horizon length of \( N = 20 \) and a time step of \
$ d_t = 0.03$\,s.
Online replanning is performed at every control loop, both on a desktop computer with an Intel i9-13900HX CPU for simulation and on the Intel NUC 13 with an i7-1360P CPU for experiment.
The neural network is built and trained using the PyTorch framework and deployed using libtorch for real-time inference.

\begin{table}[t!]
\centering
\caption{Parameters in simulation \& real world}
\resizebox{\columnwidth}{!}{
\begin{tabular}{c |c c c c c c }
    \toprule
\rowcolor{gray!15}
    \textbf{Parameters}& $diag(\mathbf{J})$ [gm$^2$] &  $m$ [kg] & $l$ [m]  & $c_{\tau}$ [1]  & $(d_x, d_y, d_z)$ [$\mathrm{s}^{-1}$] \\
\textbf{Value}& [50,53,62] & 1.55 & 0.1125   & 0.02 & (0.3,0.3,0.5) \\
    \midrule
\rowcolor{gray!15}
    \textbf{Parameters}& $\bm{\omega}_{\text{max}}$ [rad\textperiodcentered s$^{-1}$]  & $N$ & $G$ & $\mu $ & $q_c$\\
\textbf{Value} & [10,10,0.3] & 20 & 10 &500 & 30000\\
\bottomrule

\end{tabular}
}
\label{tab:param}
\vspace{-6pt}  
\end{table}

We benchmark LOONG against state-of-the-art planning frameworks for autonomous navigation. SUPER~\cite{ren2025safety}, a high-speed polynomial navigation framework, is selected for its suitability for time-critical tasks, while IPC~\cite{liu2023integrated} is included as a representative low-latency integrated planning-control baseline.
We use two open-source simulation environments
with different obstacle densities
that focus on challenging flight in \textbf{\textit{time-critical task}} (Fig.~\ref{fig:sim_highspeed}) or \textbf{\textit{cluttered forest}} (Fig.~\ref{fig:sim_forest}), 
and the LiDAR simulator provided by SUPER, which considers the limited field of view (FOV) and realistic point of LiDAR sensor~\cite{ren2025safety}. 
It should be noted that SUPER's simulator does not account for the MAV's dynamics, which assumes that SUPER's trajectories of $\bm{p}, \bm{q}, \bm{v}$ can be perfectly tracked. 
For the other two integrated planning-control frameworks, IPC uses a simplified third-order integrator assuming MPC states of $\bm{p}, \bm{v}, \bm{a}$ can be perfectly tracked
with jerk $\bm{j}$ input.
In contrast, ours incorporates an external integrator of full dynamics (\ref{eq:state_input_augmented}) with thrusts 
$\bm{f}$
as inputs.

\subsubsection{Ablation Studies in Time-Critical Task Scenario} 
\label{sec:ablation}

We first conduct an ablation study in the \textbf{\textit{time-critical task}} scenario based on the default configuration of LOONG to quantify how each component of our framework affects reliability and aggressiveness. Each configuration is evaluated over 10 trials with identical start and goal, and the success rate, average velocity, average maximum velocity, and average flight time are recorded in Table~\ref{tab:sim_ablation}, where averages are computed only over successful trials. 
The default configuration includes all modules described 
above
with a mass-normalized collective thrust limit of $f_{max}=33\text{m/s}^2$,
which
is evenly allocated to four individual thrusts as dynamic constraints in (\ref{eq:MPCC}).
Compared to the default configuration, where $G = 10$, \textbf{\textit{trajectory reuse}} is enabled for replanning, and waypoints are selected by passing through the SFC intersections, we evaluate several variants by separately modifying these components, including configurations with $G = 1$ and $G = 20$, disabling \textbf{\textit{trajectory reuse}}, and selecting waypoints without passing through the SFC intersections.
%

The results indicate that when \(G=1\), the insufficient real-time iterative solving undermines obstacle avoidance without improving flight speed. We conjecture that insufficient solver progress keeps the MPCC trajectory close to obstacles, thereby limiting the free space available for subsequent optimization. 
When \(G=20\), excessively strict and conservative constraints result in reduced aggressiveness and more solver failures.
In the absence of \textbf{\textit{trajectory reuse}}, the front-end A* may select kinetically infeasible or densely obstructed paths, which then lead to collisions or failures in the back-end MPCC.
Without passing the SFC intersection, the success rates drop substantially, since the waypoints are placed too close to obstacles, severely restricting perception range and the feasible solution space for trajectory optimization in high-speed flight.

\begin{table}[t!]
    \centering
    \caption{Ablation studies in the time-critical task.}
    \label{tab:sim_ablation}
    \sisetup{
        detect-weight, 
        mode=text,
        table-format=2.2 
    }
    \resizebox{0.49\textwidth}{!}{
    \begin{tabular}{
        ccccc
    }
    \toprule
    
    \multirow{3}{*}{\textbf{Configuration}} & \multirow{3}{*}{\makecell{\textbf{Number of} \\\textbf{Success $\uparrow$}}} & \multirow{3}{*}{\makecell{\textbf{Average} \\\textbf{Velocity (m/s) $\uparrow$}}} & \multirow{3}{*}{\makecell{\textbf{Average} \\\textbf{Maximum} \\\textbf{Velocity (m/s) $\uparrow$}}} & \multirow{3}{*}{\makecell{\textbf{Average} \\\textbf{Flight} \\\textbf{Time (s) $\downarrow$}}} 
    \\
    \\
    \\
     
    \midrule
    Default ($G=10$)& \cellcolor{blue!20}\textbf{10/10} & \cellcolor{blue!8}26.89 & \cellcolor{blue!8}42.77& \cellcolor{blue!8}3.72\\[2pt]
    $G=1$ & \cellcolor{blue!8}6/10 &26.61&40.46&3.76\\[2pt]
    $G=N=20$ & 5/10 &18.24&29.76&5.56\\[2pt]
    w.o. \textbf{\textit{Trajectory Reuse}} & 
    2/10&\cellcolor{blue!20}26.92&\cellcolor{blue!20}45.41&\cellcolor{blue!20}3.71\\[2pt]
    w.o. Passing SFC Intersection & 1/10&24.65&37.88&4.05\\[2pt]

   
    \bottomrule
    \end{tabular}
    }
    \vspace{-12pt}  
\end{table}

\subsubsection{Benchmark in Time-Critical Task Scenario} 
\label{sec:bench_highspeed}
\begin{figure}[t!]
    \centering
    \includegraphics[width=0.98\columnwidth, clip]{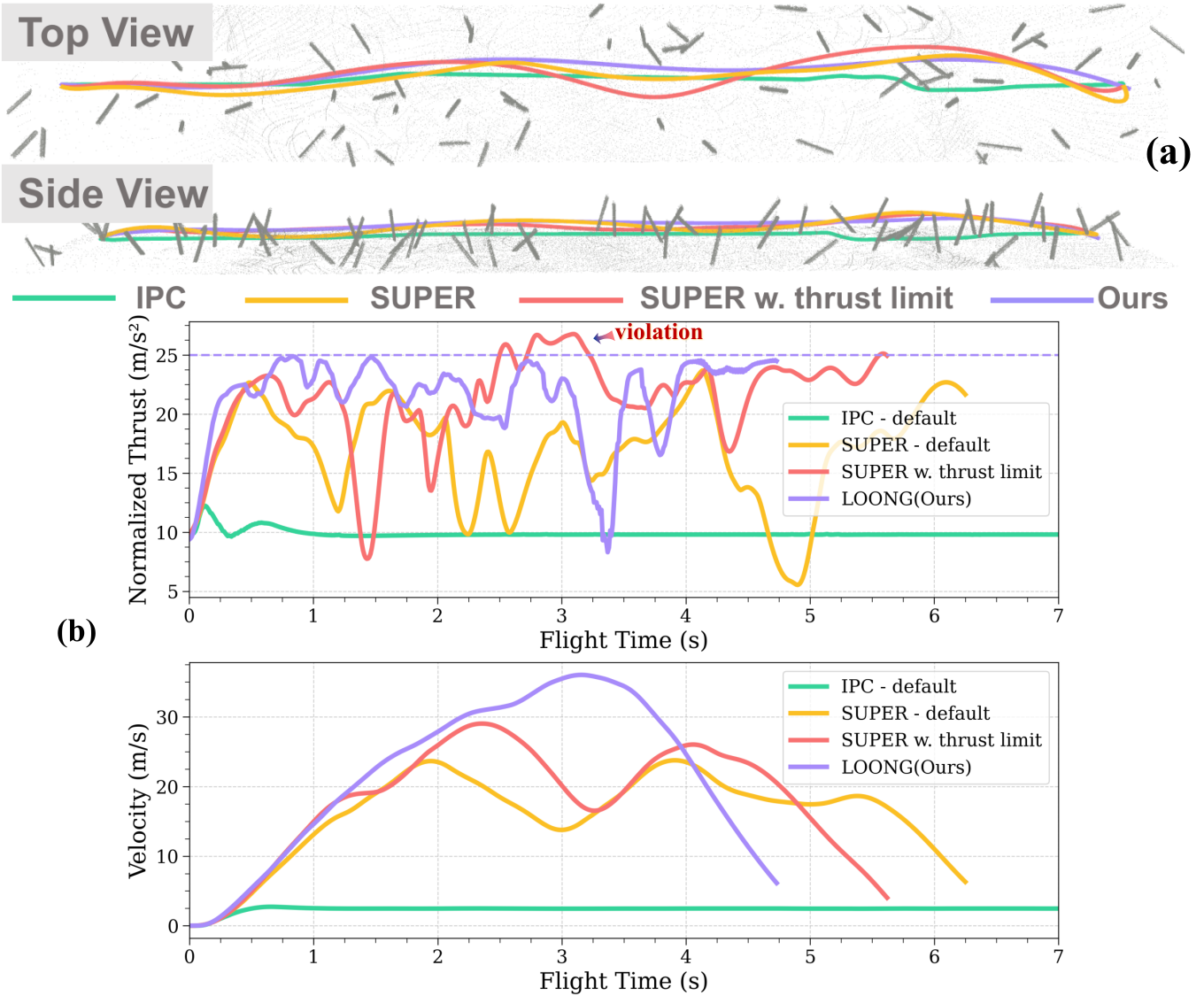}
    \caption{
    One trial of four methods with identical start and goal points. (a) Trajectories in \textbf{\textit{time-critical task}} simulation. (b) Mass-normalized collective thrust and velocity profiles. Our method fully exploits the thrust limit while maintaining obstacle avoidance.
    } 
    \vspace{-11pt}  
    \label{fig:sim_highspeed}
\end{figure}

To exhibit the effectiveness and aggressiveness of the proposed method, we benchmark LOONG against SUPER and IPC in \textbf{\textit{time-critical task}} scenario encouraging aggressive maneuvering, as shown in Fig.~\ref{fig:sim_highspeed}, 
where four configurations are evaluated in the comparison. We respect the default configurations of the baseline methods, including the default SUPER, originally designed for MAVs
with a thrust-to-weight ratio exceeding 5, and the default IPC. 
To ensure a fair comparison under identical actuator constraints, we configure both SUPER and LOONG with the same maximum collective thrust, \(f_{\max}=25\,\text{m/s}^2\). For this comparison, all other dynamic constraints in SUPER, including velocity, acceleration, and jerk limits, are removed, while the maximum collective thrust in LOONG is reduced from its default configuration to match $f_{\max}$. In addition, the default SUPER configuration is retained as a reference baseline. All methods are evaluated over 10 repeated trials.

The upper part of Table~\ref{tab:sim_comparison} records the success rates over repeated trials, computation time for planning and control, and etc. 
Since the obstacles in this scenario are relatively sparse, none of the methods experienced collisions or solver failures. 
Our method achieves greater aggressiveness in terms of average velocity, average maximum velocity, and average flight time compared to other approaches, while maintaining comparable safety.
As shown in Fig.~\ref{fig:sim_highspeed}(b),
SUPER slightly exceeds the limit \(f_{\max}=25\,\mathrm{m/s}^2\). 
We conjecture that, under a general multi-objective optimization framework, soft constraints can be violated\cite{wang2022geometrically}.
%
Despite occasionally violating the thrust limit, SUPER still exhibits longer flight times than our LOONG. We attribute the reduced aggressiveness of SUPER primarily to two factors. First, SUPER plans polynomial trajectories, whose inherent smoothness conflicts with time-optimal objectives\cite{foehn2021time}, fundamentally limiting achievable flight speed. Second, as SUPER prioritizes safety, it switches to conservative trajectories when high-speed replanning cannot be completed within the required time, which further reduces flight aggressiveness.
In contrast, our method can reach the input limits without violation, ensuring that dynamic constraints are satisfied even when the MAV is operating at actuator limits, while achieving a replanning frequency of 100\,Hz,
aligning with IPC in planning-control integration efficiency. Moreover, although IPC
supports high planning frequency, it is overly conservative due to its naive average time allocation,
and is therefore excluded from subsequent benchmarks.

\begin{table*}[ht]
    \centering
    \caption{Comparison of different framework in Simulation.}
    \label{tab:sim_comparison}
    \sisetup{
        detect-weight, 
        mode=text,
        table-format=2.2 
    }
    \resizebox{0.9\textwidth}{!}{
    \begin{tabular}{
        ccccccc
    }
    \toprule
    
    \multirow{3}{*}{\textbf{Scenario}} & \multirow{3}{*}{\textbf{Planning Framework}} & \multirow{3}{*}{\makecell{\textbf{Number of} \\\textbf{Success $\uparrow$}}} & \multirow{3}{*}{\makecell{\textbf{Average} \\\textbf{Velocity (m/s) $\uparrow$}}} & \multirow{3}{*}{\makecell{\textbf{Average} \\\textbf{Maximum} \\\textbf{Velocity (m/s) $\uparrow$}}} & \multirow{3}{*}{\makecell{\textbf{Average} \\\textbf{Computation} \\\textbf{Time (ms) $\downarrow$}}} & \multirow{3}{*}{\makecell{\textbf{Average} \\\textbf{Flight} \\\textbf{Time (s) $\downarrow$}}} 
    \\
    \\
    \\
     
    \midrule

   
   



\multirow{4}{*}{\makecell{\textbf{\textit{time-critical task}} \\ $15\mathrm{m}\times120\mathrm{m}\times4\mathrm{m}$}}
 & \makecell{SUPER - default}
   & \cellcolor{blue!20}\textbf{10/10} & 16.23 & 24.88 & 24.86 & 6.24\\

 & IPC - default
   & \cellcolor{blue!20}\textbf{10/10} & 2.85 & 3.19 & \cellcolor{blue!20}\textbf{2.41} & 40.57\\
   
 & SUPER - w. only thrust limit
   & \cellcolor{blue!20}\textbf{10/10} & \cellcolor{blue!8}17.99 & \cellcolor{blue!8}29.91 & 25.28 & \cellcolor{blue!8}5.62\\

 & LOONG(Ours)
   & \cellcolor{blue!20}\textbf{10/10}
   & \cellcolor{blue!20}\textbf{22.10}
   & \cellcolor{blue!20}\textbf{33.15}
   & \cellcolor{blue!8}5.87
   & \cellcolor{blue!20}\textbf{4.64}\\
   
\midrule

\multirow{4}{*}{\makecell{\textbf{\textit{cluttered forest}}\\ $60\mathrm{m}\times60\mathrm{m}\times5\mathrm{m}$}}
 & SUPER - \textit{high-speed} 
   &4/10 &\cellcolor{blue!8}7.34 &\cellcolor{blue!8}17.63 &47.98 &\cellcolor{blue!8}8.81\\
 & LOONG(Ours) - \textit{high-speed}
   & \cellcolor{blue!8}6/10
   & \cellcolor{blue!20}\textbf{13.63}
   & \cellcolor{blue!20}\textbf{20.60}
   & \cellcolor{blue!8}6.84
   & \cellcolor{blue!20}\textbf{4.56}\\
 & SUPER - \textit{dense}
   & \cellcolor{blue!20}\textbf{10/10} &6.81 &8.00 &50.11 &9.53\\

 & LOONG(Ours) - \textit{dense}
   & \cellcolor{blue!20}\textbf{10/10}
   & 6.94
   & 9.28
   & \cellcolor{blue!20}\textbf{6.43}
   & 9.15\\

    \bottomrule
    \end{tabular}
    }
    \vspace{-12pt}
\end{table*}

\begin{figure}[t!]
    \centering
    \includegraphics[width=0.98\columnwidth, clip]{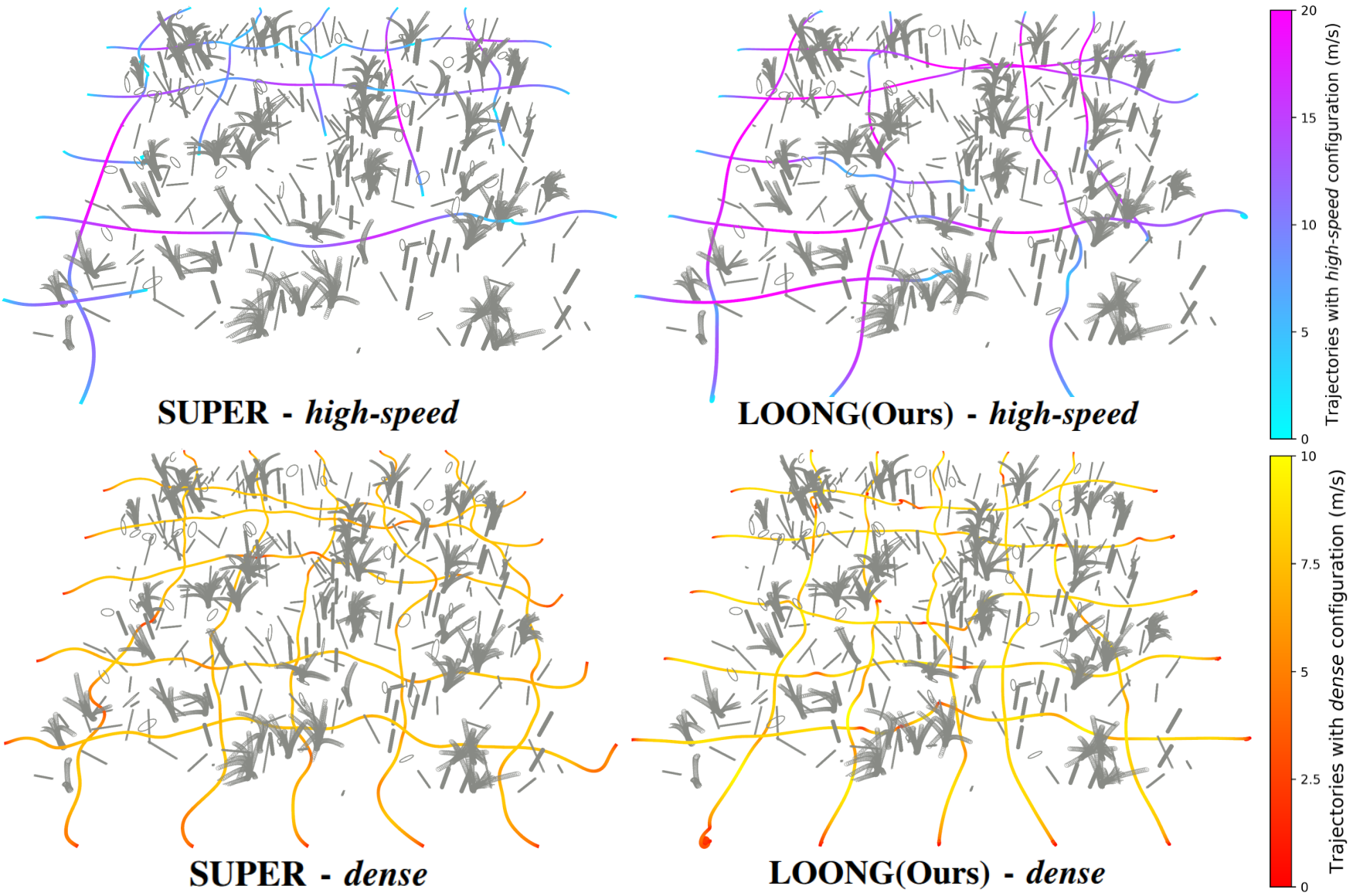}
    \caption{Trajectories of four planning configurations starting from 10 different points in the \textbf{\textit{cluttered forest}} scenario. Some \textit{high-speed} trials failed while all \textit{dense} trials succeeded. Ours achieved a more agile performance compared to SUPER.}
    \vspace{-12pt}  
    \label{fig:sim_forest}
\end{figure}

\subsubsection{Benchmark in Cluttered Forest} 

To evaluate the robustness of LOONG, we conduct experiments in the \textbf{\textit{cluttered forest}} scenario and compare its performance with SUPER. For a fair comparison against SUPER's two open-source configurations \textit{high-speed} (default) and \textit{dense}, the velocity limit of LOONG, ${v}_{\theta,max}$, is reduced from 
its default configuration to 25\,m/s or 8\,m/s, respectively, to match SUPER's maximum speeds $\bm{v}$. 
For each configuration, 10 trials are conducted with different start and goal points. 
The experimental results are summarized in the lower part of Table III and Fig.~\ref{fig:sim_forest}.
Table~\ref{tab:sim_comparison} indicates that for \textit{high-speed} configuration, LOONG achieves faster flights and higher success rates by leveraging actuators more sufficiently and maintaining a higher replanning frequency. 
For \textit{dense} configuration, LOONG finishes all flights
while demonstrating superior agility and computational efficiency.
Notably, 
compared to Section\,\ref{sec:bench_highspeed},
while SUPER manages \textbf{\textit{time-critical task}} by \textit{high-speed} configuration and \textbf{\textit{cluttered forest}} by \textit{dense} configuration, it necessitates extensive parameter tuning, such as planning horizon, penalty weights, resolution,
etc., whereas LOONG requires only one parameter ${v}_{\theta,max}$ adjustment.
Moreover,
due to relatively stable MPCC complexity and consistent neural network inference time, LOONG's computation time is largely insensitive to environmental complexity, 5.87\,ms and 6.43\,ms respectively, 
whereas SUPER's computation time nearly doubles from 24.86\,ms in \textbf{\textit{time-critical task}} to 50.11\,ms in \textbf{\textit{cluttered forest}}.

\vspace{-2pt}\subsection{Real-world Experiments}

\subsubsection{Real-world Experiment Setup} 

\begin{figure}[t!] 
    \centering
    \includegraphics[width=0.4\textwidth]{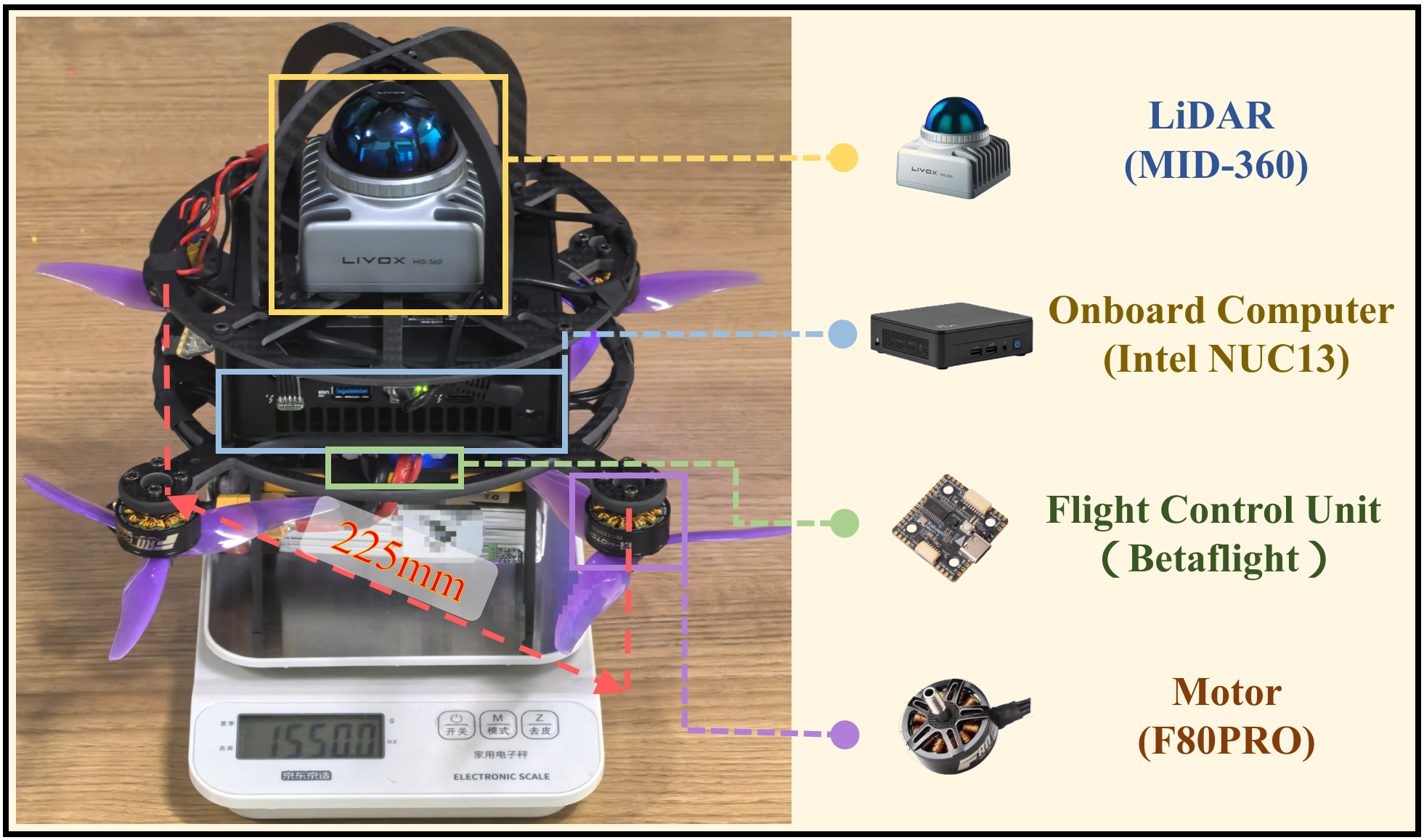} 
    \caption{Hardware platform of LOONG.}
    \label{fig:platform}
    \vspace{-12pt}
\end{figure}

We validate the system's effectiveness
and reliability through real-world experiments.
The experiments can be found in the [\href{https://youtu.be/vexXXhv99oQ}{\textbf{\emph{video}}}].
For the hardware platform, we deploy a custom-built compact MAV with a 225\,mm wheelbase,
as shown in Fig.~\ref{fig:platform}.
The fully loaded 
vehicle 
weighs 1.5\,kg and delivers a thrust-to-weight ratio exceeding 5.0. It is actuated by T-Motor F80PRO motors, which are typically used in racing drone,
providing high thrust (up to 21.1 N with 5-inch propellers) and rapid response.
The onboard computation is performed by an Intel NUC 13 computer featuring a 12-core 5.0\,GHz CPU, and a flight control unit (FCU) running Betaflight with custom firmware. For perception, the MAV is equipped with a MID-360 3D LiDAR sensor. 
FAST-LIO2 \cite{xu2022FASTLIO2} 
is used
to fuse data from the LiDAR and IMU sensors, while a sliding point-cloud map~\cite{ren2025safety} is used to represent the occupied spaces of the environment.
For planning and control, the default configuration of LOONG is used.

\subsubsection{High-Speed Autonomous Flights in Clutter}

\begin{figure}[t!]
    \centering
    \includegraphics[width=0.49\textwidth, clip]{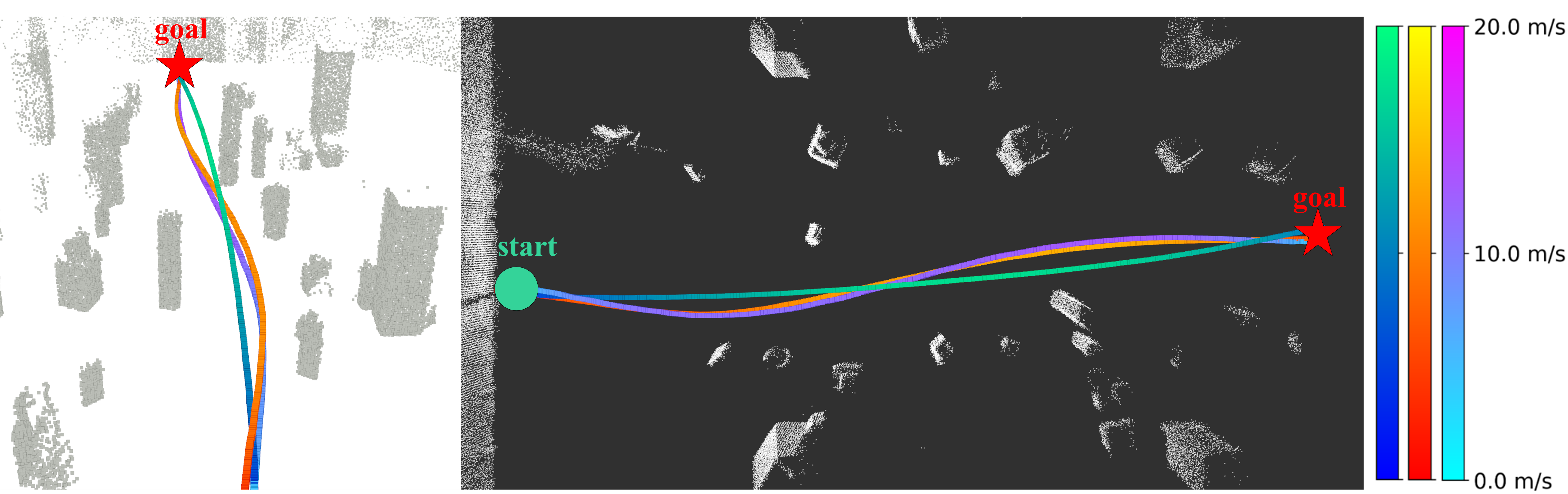}
    \caption{
    Real-world trajectories visualization for repeated high-speed flights,
    demonstrating the reproducible aggressive behavior.
    }
    \vspace{-11pt}  
    \label{fig:real_highspeed_repeated}
\end{figure}

To test the systematic effectiveness and aggressiveness of the proposed method,
we first have our MAV perform an aggressive flight in a 20\,m indoor scenario with obstacles.
As shown in Fig.~\ref{fig:real_highspeed}, 
the MAV begins at a hover height of $1.5\,\text{m}$ at $t=0\,\text{s}$ and traverses a 
$10\,\text{m}\times 10\,\text{m}$ 
cluttered region to reach the goal. 
The obstacle field has a density of 0.30 obstacles$/\mathrm{m}^2$, with obstacle heights in the range $[0.8\,\mathrm{m}, 3.0\,\mathrm{m}]$ and obstacle radii in the range $[0.2\,\mathrm{m}, 0.6\,\mathrm{m}]$.
The maximum gap between adjacent obstacles is about 2.5\,m.
As shown in Fig.~\ref{fig:real_highspeed}(b), the velocity profile indicates that the MAV reaches 18\,m/s in about 1.2\,s, while the throttle profile normalized by the collective thrust limit shows that the system can operate at its maximum thrust, enabling agile maneuvering among obstacles during the real-time autonomous flight.
Three additional experiments are conducted to evaluate the robustness of the proposed system, as shown in Fig. 10.

\subsubsection{Consecutive Trials with Random Start Points} 

\begin{figure*}[t!]
    \centering
    \includegraphics[width=0.95\textwidth, clip]{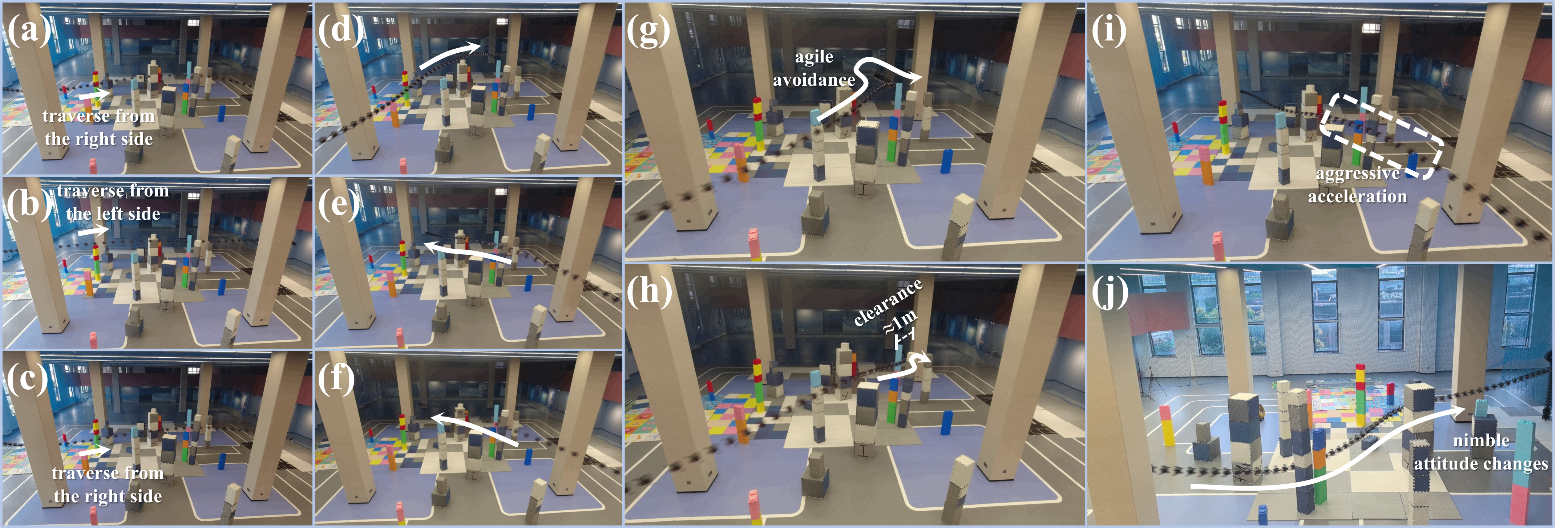}
    \caption{
    Real-world trajectories from ten consecutive trials with random start points. Snapshots in the composite trajectory plot are taken at 0.033\,s intervals.  
    The MAV autonomously performs various obstacle avoidance behaviors:  
    (a)-(c) bypassing obstacles at high speed from different sides,  
    (d)-(f) gaining altitude to avoid densely cluttered regions,  
    (g) aggressive avoidance maneuvers in response to close obstacles,  
    (h) rapid traversal through a narrow gap,  
    (i) large pitch maneuver for acceleration,  
    (j) nimble attitude changes.
    }
    \vspace{-12pt}  
    \label{fig:real_random}
\end{figure*}

\begin{figure}[t!]
    \centering
    \includegraphics[width=\columnwidth, clip]{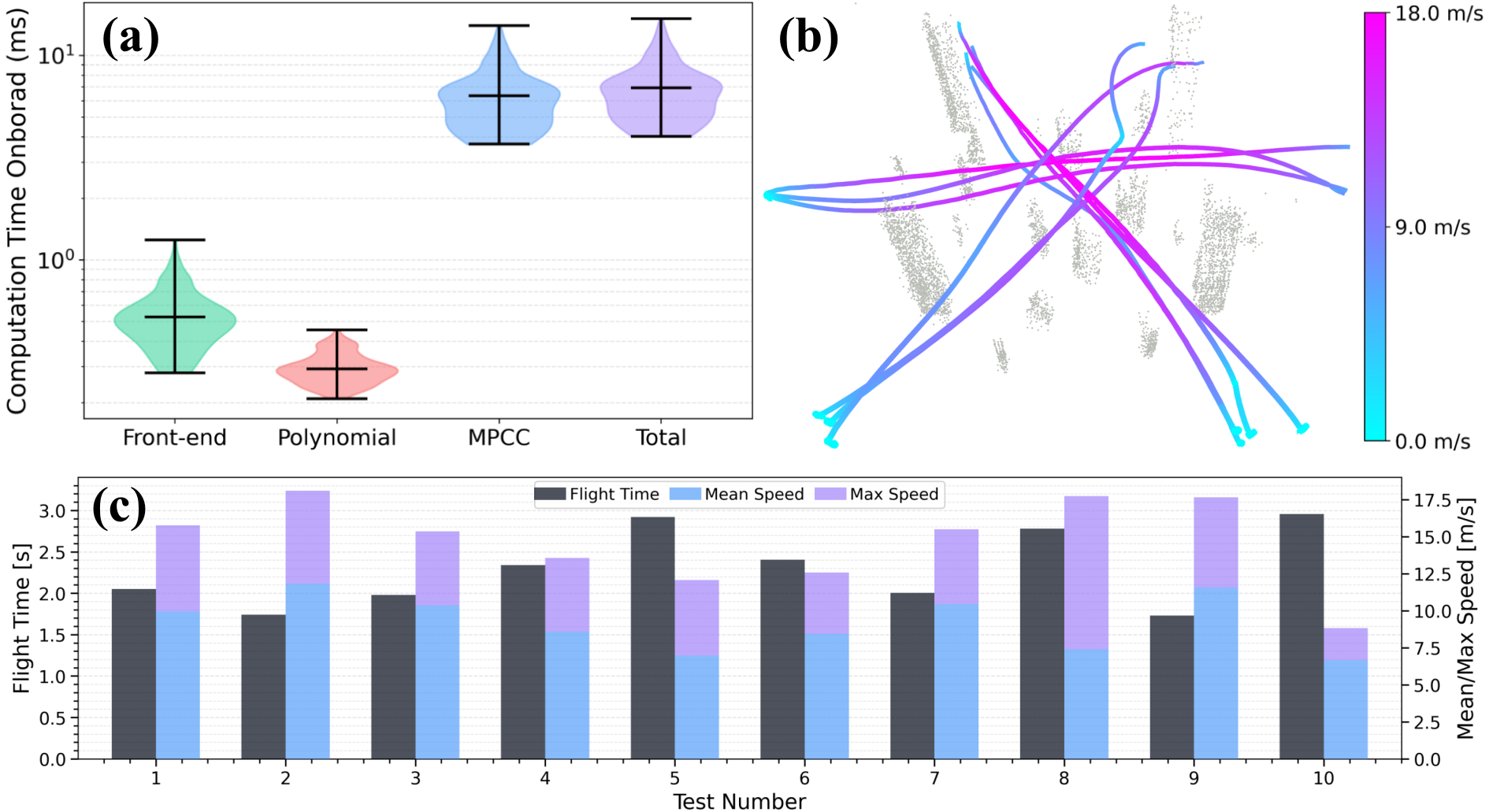}
    \caption{
    Real-world consecutive trial data of LOONG. (a) Onboard computation time distribution per step, with total time below 10\,ms to achieve 100\,Hz replanning. (b) Visualization of ten trajectories.
    (c) Statistics of flight time and velocity.
    }
    \vspace{-12pt}  
    \label{fig:real_random_data}
\end{figure}

As shown in Fig.~\ref{fig:real_random}, the MAV adaptively maneuvered through free space between obstacles in ten consecutive trials of 18\,m forward flight, demonstrating the real-world generalization and reliability of our system with the default LOONG configuration. 
The MAV traversed from three directions, with initial positions randomly selected within a 2\,m radius and initial yaw angles within \(\pm 10^\circ\).
Fig.~\ref{fig:real_random_data} presents the onboard computation time of each planning component per loop.
It can be observed that the proposed path planning and reference generation methods require approximately 0.5\,ms and 0.3\,ms per loop, respectively, demonstrating high computational efficiency. By significantly reducing the computational burden of the front-end modules, sufficient onboard computational resources are reserved for the more time-consuming MPCC, thereby enabling the MAV to achieve online planning at a frequency of 100 Hz.
Despite traversing highly cluttered obstacles, all ten consecutive flights succeed, as depicted in Fig.~\ref{fig:real_random_data}(b), and achieved an average peak speed of 14.72 m/s and an average flight time of 2.29 s, as illustrated in Fig.~\ref{fig:real_random_data}(c). 
These results strongly demonstrate our framework's capability to autonomously generate safe and agile trajectories in challenging environments in real time.

\section{Conclusion and Discussion}
This article presents a learning-accelerated online time-optimal planning and control framework termed LOONG for MAVs' high-speed autonomous flight, 
which exhibits both aggressive flight performance and safety in simulation and real-world experiments.
Several improvements are envisioned for future work.  
For the optimization scheme, while the current RTI solver enables real-time performance, its convergence may be challenged in extremely dynamic environments. 
The recent advancement of onboard GPUs and the sampling-based scheme of model predictive path integral 
(MPPI)
could overcome this bottleneck by enabling massive parallel roll-outs in real time.
Moreover, our benchmark demonstrates LOONG's 
generalization
by merely adjusting the maximum progress speed. Thereby, environment-adaptive tuning remains a promising enhancement. 
Finally, given LOONG's fully autonomous design and generalization, expanding its application to outdoor and field environments offers an exciting frontier for achieving even greater performance.  







\bibliographystyle{Bibliography/IEEEtranTIE}
\bibliography{Bibliography/IEEEabrv,Bibliography/BIB_xx-TIE-xxxx,Bibliography/reference_TIE}\ 

@inproceedings{mellinger2011minimum,
  title={Minimum snap trajectory generation and control for quadrotors},
  author={Mellinger, Daniel and Kumar, Vijay},
  booktitle={2011 IEEE International Conference on Robotics and Automation},
  pages={2520--2525},
  year={2011},
  organization={IEEE}
}

@inproceedings{richter2016polynomial,
  title={Polynomial trajectory planning for aggressive quadrotor flight in dense indoor environments},
  author={Richter, Charles and Bry, Adam and Roy, Nicholas},
  booktitle={Robotics Research: The 16th International Symposium ISRR},
  pages={649--666},
  year={2016},
  organization={Springer}
}

@article{faessler2017differential,
  title={Differential flatness of quadrotor dynamics subject to rotor drag for accurate tracking of high-speed trajectories},
  author={Faessler, Matthias and Franchi, Antonio and Scaramuzza, Davide},
  journal={IEEE Robotics and Automation Letters},
  volume={3},
  number={2},
  pages={620--626},
  year={2017},
  publisher={IEEE}
}

@article{moon2019challenges,
  title={Challenges and implemented technologies used in autonomous drone racing},
  author={Moon, Hyungpil and Martinez-Carranza, Jose and Cieslewski, Titus and Faessler, Matthias and Falanga, Davide and Simovic, Alessandro and Scaramuzza, Davide and Li, Shuo and Ozo, Michael and De Wagter, Christophe and others},
  journal={Intelligent Service Robotics},
  pages={1--12},
  year={2019},
  publisher={Springer}
}

@article{foehn2021time,
  title={Time-optimal planning for quadrotor waypoint flight},
  author={Foehn, Philipp and Romero, Angel and Scaramuzza, Davide},
  journal={Science Robotics},
  volume={6},
  number={56},
  pages={eabh1221},
  year={2021},
  publisher={American Association for the Advancement of Science}
}

@article{zhou2022swarm,
  title={Swarm of micro flying robots in the wild},
  author={Zhou, Xin and Wen, Xiangyong and Wang, Zhepei and Gao, Yuman and Li, Haojia and Wang, Qianhao and Yang, Tiankai and Lu, Haojian and Cao, Yanjun and Xu, Chao and others},
  journal={Science Robotics},
  volume={7},
  number={66},
  pages={eabm5954},
  year={2022},
  publisher={American Association for the Advancement of Science}
}

@article{romero2022time,
  title={Time-optimal online replanning for agile quadrotor flight},
  author={Romero, Angel and Penicka, Robert and Scaramuzza, Davide},
  journal={IEEE Robotics and Automation Letters},
  volume={7},
  number={3},
  pages={7730--7737},
  year={2022},
  publisher={IEEE}
}

@article{romero2022model,
  title={Model predictive contouring control for time-optimal quadrotor flight},
  author={Romero, Angel and Sun, Sihao and Foehn, Philipp and Scaramuzza, Davide},
  journal={IEEE Transactions on Robotics},
  volume={38},
  number={6},
  pages={3340--3356},
  year={2022},
  publisher={IEEE}
}

@article{loquercio2021learning,
  title={Learning high-speed flight in the wild},
  author={Loquercio, Antonio and Kaufmann, Elia and Ranftl, Ren{\'e} and M{\"u}ller, Matthias and Koltun, Vladlen and Scaramuzza, Davide},
  journal={Science Robotics},
  volume={6},
  number={59},
  pages={eabg5810},
  year={2021},
  publisher={American Association for the Advancement of Science}
}

@article{kaufmann2023champion,
  title={Champion-level drone racing using deep reinforcement learning},
  author={Kaufmann, Elia and Bauersfeld, Leonard and Loquercio, Antonio and M{\"u}ller, Matthias and Koltun, Vladlen and Scaramuzza, Davide},
  journal={Nature},
  volume={620},
  number={7976},
  pages={982--987},
  year={2023},
  publisher={Nature Publishing Group UK London}
}

@article{wang2022geometrically,
  title={Geometrically constrained trajectory optimization for multicopters},
  author={Wang, Zhepei and Zhou, Xin and Xu, Chao and Gao, Fei},
  journal={IEEE Transactions on Robotics},
  volume={38},
  number={5},
  pages={3259--3278},
  year={2022},
  publisher={IEEE}
}

@article{verschueren2022acados,
  title={acados-a modular open-source framework for fast embedded optimal control},
  author={Verschueren, Robin and Frison, Gianluca and Kouzoupis, Dimitris and Frey, Jonathan and Duijkeren, Niels van and Zanelli, Andrea and Novoselnik, Branimir and Albin, Thivaharan and Quirynen, Rien and Diehl, Moritz},
  journal={Mathematical Programming Computation},
  volume={14},
  number={1},
  pages={147--183},
  year={2022},
  publisher={Springer}
}

@article{hanover2024autonomous,
  title={Autonomous drone racing: A survey},
  author={Hanover, Drew and Loquercio, Antonio and Bauersfeld, Leonard and Romero, Angel and Penicka, Robert and Song, Yunlong and Cioffi, Giovanni and Kaufmann, Elia and Scaramuzza, Davide},
  journal={IEEE Transactions on Robotics},
  year={2024},
  publisher={IEEE}
}

@inproceedings{guan2025learning,
  title={Learning Time-Optimal Online Replanning for Distributed Model Predictive Contouring Control of Quadrotors},
  author={Guan, Xin and Zhao, Fangguo and Tian, Shunxin and Li, Shuo},
  booktitle={2025 IEEE International Conference on Robotics and Automation (ICRA)},
  pages={14527--14533},
  year={2025},
  organization={IEEE}
}

@article{mohta2018fast,
  title={Fast, autonomous flight in GPS-denied and cluttered environments},
  author={Mohta, Kartik and Watterson, Michael and Mulgaonkar, Yash and Liu, Sikang and Qu, Chao and Makineni, Anurag and Saulnier, Kelsey and Sun, Ke and Zhu, Alex and Delmerico, Jeffrey and others},
  journal={Journal of Field Robotics},
  volume={35},
  number={1},
  pages={101--120},
  year={2018},
  publisher={Wiley Online Library}
}

@article{mishra2020drone,
  title={Drone-surveillance for search and rescue in natural disaster},
  author={Mishra, Balmukund and Garg, Deepak and Narang, Pratik and Mishra, Vipul},
  journal={Computer Communications},
  volume={156},
  pages={1--10},
  year={2020},
  publisher={Elsevier}
}

@article{daud2022applications,
  title={Applications of drone in disaster management: A scoping review},
  author={Daud, Sharifah Mastura Syed Mohd and Yusof, Mohd Yusmiaidil Putera Mohd and Heo, Chong Chin and Khoo, Lay See and Singh, Mansharan Kaur Chainchel and Mahmood, Mohd Shah and Nawawi, Hapizah},
  journal={Science \& Justice},
  volume={62},
  number={1},
  pages={30--42},
  year={2022},
  publisher={Elsevier}
}

@article{song2023reaching,
  title={Reaching the limit in autonomous racing: Optimal control versus reinforcement learning},
  author={Song, Yunlong and Romero, Angel and M{\"u}ller, Matthias and Koltun, Vladlen and Scaramuzza, Davide},
  journal={Science Robotics},
  volume={8},
  number={82},
  pages={eadg1462},
  year={2023},
  publisher={American Association for the Advancement of Science}
}

@article{liu2023integrated,
  title={Integrated planning and control for quadrotor navigation in presence of suddenly appearing objects and disturbances},
  author={Liu, Wenyi and Ren, Yunfan and Zhang, Fu},
  journal={IEEE Robotics and Automation Letters},
  volume={9},
  number={1},
  pages={899--906},
  year={2023},
  publisher={IEEE}
}

@article{zhou2019robust,
  title={Robust and efficient quadrotor trajectory generation for fast autonomous flight},
  author={Zhou, Boyu and Gao, Fei and Wang, Luqi and Liu, Chuhao and Shen, Shaojie},
  journal={IEEE Robotics and Automation Letters},
  volume={4},
  number={4},
  pages={3529--3536},
  year={2019},
  publisher={IEEE}
}

@article{wang2025fast,
  title={Fast iterative region inflation for computing large 2-D/3-D convex regions of obstacle-free space},
  author={Wang, Qianhao and Wang, Zhepei and Wang, Mingyang and Ji, Jialin and Han, Zhichao and Wu, Tianyue and Jin, Rui and Gao, Yuman and Xu, Chao and Gao, Fei},
  journal={IEEE Transactions on Robotics},
  year={2025},
  publisher={IEEE}
}

@inproceedings{yuan2025safety,
  title={Safety-Critical Online Quadrotor Trajectory Planner for Agile Flights in Unknown Environments},
  author={Yuan, Jiazhe and Cao, Dongcheng and Mei, Jiahao and Chen, Jiming and Li, Shuo},
  booktitle={2025 IEEE International Conference on Robotics and Automation (ICRA)},
  pages={11773--11779},
  year={2025},
  organization={IEEE}
}

@article{ren2025safety,
  title={Safety-assured high-speed navigation for MAVs},
  author={Ren, Yunfan and Zhu, Fangcheng and Lu, Guozheng and Cai, Yixi and Yin, Longji and Kong, Fanze and Lin, Jiarong and Chen, Nan and Zhang, Fu},
  journal={Science Robotics},
  volume={10},
  number={98},
  pages={eado6187},
  year={2025},
  publisher={American Association for the Advancement of Science}
}

@article{zhang2024back,
  title={Back to Newton's Laws: Learning Vision-based Agile Flight via Differentiable Physics},
  author={Zhang, Yuang and Hu, Yu and Song, Yunlong and Zou, Danping and Lin, Weiyao},
  journal={arXiv preprint arXiv:2407.10648},
  year={2024}
}

@ARTICLE{lu2023On,
  author={Lu, Guozheng and Xu, Wei and Zhang, Fu},
  journal={IEEE Transactions on Industrial Electronics}, 
  title={On-Manifold Model Predictive Control for Trajectory Tracking on Robotic Systems}, 
  year={2023},
  volume={70},
  number={9},
  pages={9192-9202},
  doi={10.1109/TIE.2022.3212397}}

@ARTICLE{9143458,
  author={Zhou, Yu and Lai, Shupeng and Cheng, Huimin and Redhwan, Abdul Hamid Mohamed and Wang, Pengfei and Zhu, Junji and Gao, Zhi and Ma, Zhengtian and Bi, Yingcai and Lin, Feng and Chen, Ben M.},
  journal={IEEE Transactions on Industrial Electronics}, 
  title={Toward Autonomy of Micro Aerial Vehicles in Unknown and Global Positioning System Denied Environments}, 
  year={2021},
  volume={68},
  number={8},
  pages={7642-7651},
  doi={10.1109/TIE.2020.3008378}}

@ARTICLE{10649014,
  author={Li, Ruocheng and Xin, Bin},
  journal={IEEE Transactions on Industrial Electronics}, 
  title={Autonomous Navigation of Quadrotors in Dynamic Complex Environments}, 
  year={2025},
  volume={72},
  number={3},
  pages={2790-2800},
  doi={10.1109/TIE.2024.3433585}}

@ARTICLE{9900135,
  author={Liu, Kangcheng and Chen, Ben M.},
  journal={IEEE Transactions on Industrial Electronics}, 
  title={Industrial UAV-Based Unsupervised Domain Adaptive Crack Recognitions: From Database Towards Real-Site Infrastructural Inspections}, 
  year={2023},
  volume={70},
  number={9},
  pages={9410-9420},
  doi={10.1109/TIE.2022.3204953}}

@ARTICLE{xu2022FASTLIO2,
  author={Xu, Wei and Cai, Yixi and He, Dongjiao and Lin, Jiarong and Zhang, Fu},
  journal={IEEE Transactions on Robotics}, 
  title={FAST-LIO2: Fast Direct LiDAR-Inertial Odometry}, 
  year={2022},
  volume={38},
  number={4},
  pages={2053-2073},
  doi={10.1109/TRO.2022.3141876}}

@article{KOHLER2024100929,
title = {Analysis and design of model predictive control frameworks for dynamic operation-An overview},
journal = {Annual Reviews in Control},
volume = {57},
pages = {100929},
year = {2024},
issn = {1367-5788},
doi = {https://doi.org/10.1016/j.arcontrol.2023.100929},
author = {Johannes Kohler and Matthias A. Muller and Frank Allgower}
}

@inproceedings{oleynikova2016continuous,
  title={Continuous-time trajectory optimization for online uav replanning},
  author={Oleynikova, Helen and Burri, Michael and Taylor, Zachary and Nieto, Juan and Siegwart, Roland and Galceran, Enric},
  booktitle={2016 IEEE/RSJ international conference on intelligent robots and systems (IROS)},
  pages={5332--5339},
  year={2016},
  organization={IEEE}
}

@inproceedings{oleynikova2017voxblox,
  title={Voxblox: Incremental 3d euclidean signed distance fields for on-board mav planning},
  author={Oleynikova, Helen and Taylor, Zachary and Fehr, Marius and Siegwart, Roland and Nieto, Juan},
  booktitle={2017 IEEE/RSJ International Conference on Intelligent Robots and Systems (IROS)},
  pages={1366--1373},
  year={2017},
  organization={IEEE}
}

@article{liu2017planning,
  title={Planning dynamically feasible trajectories for quadrotors using safe flight corridors in 3-d complex environments},
  author={Liu, Sikang and Watterson, Michael and Mohta, Kartik and Sun, Ke and Bhattacharya, Subhrajit and Taylor, Camillo J and Kumar, Vijay},
  journal={IEEE Robotics and Automation Letters},
  volume={2},
  number={3},
  pages={1688--1695},
  year={2017},
  publisher={IEEE}
}

@inproceedings{ren2022bubble,
  title={Bubble planner: Planning high-speed smooth quadrotor trajectories using receding corridors},
  author={Ren, Yunfan and Zhu, Fangcheng and Liu, Wenyi and Wang, Zhepei and Lin, Yi and Gao, Fei and Zhang, Fu},
  booktitle={2022 IEEE/RSJ International Conference on Intelligent Robots and Systems (IROS)},
  pages={6332--6339},
  year={2022},
  organization={IEEE}
}

\end{document}